%% file: main.tex
\pdfoutput=1

\documentclass[11pt]{article}

\usepackage[final]{acl}

\usepackage{times}
\usepackage{latexsym}

\usepackage[T1]{fontenc}

\usepackage[utf8]{inputenc}

\usepackage{microtype}

\usepackage{inconsolata}

\usepackage{graphicx}

\usepackage{amsthm,amsmath,amssymb,amsfonts}
\usepackage{wrapfig}
\usepackage{caption}
\usepackage{subcaption}
\usepackage{booktabs}       
\usepackage{xcolor,colortbl}
\usepackage[most]{tcolorbox}
\usepackage{multirow}
\usepackage{arydshln}
\usepackage{stackengine}  
\usepackage{soul} 
\usepackage{listings}
\usepackage{fvextra}

\usepackage{xspace}

\newcommand{\llm}[0]{LLM\xspace}
\newcommand{\llms}[0]{LLMs\xspace}
\newcommand{\mds}[0]{MDS\xspace}
\newcommand{\md}[0]{MDS\xspace}
\newcommand{\eg}[0]{\textit{e.g.,}\xspace}
\newcommand{\ie}[0]{\textit{i.e.,}\xspace}

\newcommand{\newsdataset}[0]{SummHay-News\xspace}
\newcommand{\convdataset}[0]{SummHay-Conv\xspace}
\newcommand{\postspace}{\vskip -3mm}
\newcommand{\minipostspace}{\vskip -1.5mm}

\newcommand{\fullcov}[0]{\texttt{FULL}\xspace}
\newcommand{\partcov}[0]{\texttt{PARTIAL}\xspace}
\newcommand{\nocov}[0]{\texttt{NO}\xspace}

\newcommand{\subtopic}[0]{``subtopic''\xspace}
\newcommand{\subtopictrust}[0]{``subtopic+trustworthy''\xspace}

\newcommand{\fullchatgpt}[0]{\texttt{gpt-3.5-turbo-0125}\xspace}
\newcommand{\fullgptmini}[0]{\texttt{gpt-4o-mini-2024-07-18}\xspace}
\newcommand{\fullgptfour}[0]{\texttt{gpt-4o-2024-05-13}\xspace}
\newcommand{\fullllama}[0]{\texttt{llama-v3p1-70b-instruct}\xspace}
\newcommand{\fullqwen}[0]{\texttt{qwen2-72b-instruct}\xspace}
\newcommand{\fullgemini}[0]{\texttt{gemini-1.5-flash}\xspace}

\newcommand{\shortchatgpt}[0]{\texttt{GPT-3.5-Turbo}\xspace}

\newcommand{\shortgptfour}[0]{\texttt{GPT-4o}\xspace}
\newcommand{\shortllama}[0]{\texttt{Llama 3.1 (70B)}\xspace}
\newcommand{\shortqwen}[0]{\texttt{Qwen 2 (72B)}\xspace}
\newcommand{\shortgemini}[0]{\texttt{Gemini (Flash)}\xspace}

\title{
From Single to Multi:\\How LLMs Hallucinate in Multi-Document Summarization
}

\author{Catarina G. Belem\thanks{~Work done while interning at Megagon Labs.}\\
  University of California Irvine \\
  \texttt{cbelem@uci.edu} \\\And
  Pouya Pezeshkpour \\
  Megagon Labs \\
  \texttt{pouya@megagon.ai}\\\And
  Hayate Iso \\
  Megagon Labs \\
  \texttt{hayate@megagon.ai}\\
  \AND
  Seiji Maekawa \\
  Megagon Labs \\
  \texttt{seiji@megagon.ai} \\\And
  Nikita Bhutani \\ 
  Megagon Labs \\
  \texttt{nikita@megagon.ai} \\\And
  Estevam Hruschka \\ 
  Megagon Labs \\
  \texttt{estevam@megagon.ai} 
}

\begin{document}
\maketitle
\begin{abstract}
Although many studies have investigated and reduced hallucinations in large language models (LLMs) for single-document tasks, research on hallucination in multi-document summarization (MDS) tasks remains largely unexplored.
Specifically, it is unclear how the challenges arising from handling multiple documents (\textit{e.g.}, repetition and diversity of information) affect models outputs.
In this work, we investigate how hallucinations manifest in LLMs when summarizing topic-specific information from a set of documents.
Since no benchmarks exist for investigating hallucinations in MDS, we leverage existing news and conversation datasets, annotated with topic-specific insights, to create two novel multi-document benchmarks. 
When evaluating 5 LLMs on our benchmarks, we observe that on average, up to 75\% of the content in LLM-generated summary is hallucinated, with hallucinations more likely to occur towards the end of the summaries.
Moreover, when summarizing non-existent topic-related information, \texttt{gpt-3.5-turbo} and \texttt{GPT-4o} still generate summaries about 79.35\% and 44\% of the time, raising concerns about their tendency to fabricate content.
To better understand the characteristics of these hallucinations, we conduct a human evaluation of 700+ insights and discover that most errors stem from either failing to follow instructions or producing overly generic insights. 
Motivated by these observations, we investigate the efficacy of simple \textit{post-hoc} baselines in mitigating hallucinations but find them only moderately effective. 
Our results underscore the need for more effective approaches to systematically mitigate hallucinations in MDS. 
\end{abstract}
%
%
\section{Introduction}
\label{sec:introduction}
Multi-document summarization (MDS) has numerous real-world applications, including planning treatments and diagnosing patients based on their medical history (\ie doctor notes, lab reports)~\citep{Tang2023}, forming legal arguments by linking precedents~\citep{Rodgers2023-how-techn-is-transf-law-firms,wu-etal-2023-precedent}, or screening resumes to match candidates to job descriptions~\citep{wang2024jobfairframeworkbenchmarkinggender,Du2024}. 
These tasks often involve linking information across lengthy documents, making the summarization process time- and labor-intensive~\citep{VanVeen2024}. 
Recently, large language models (\llms) have been proposed as more efficient approaches to reduce the human effort and increase scalability~\citep{Katz2023,VanVeen2024,Liu2024-Exploring-potential-medical}.
\begin{figure}[tb]
    \centering
    \includegraphics[width=\columnwidth]{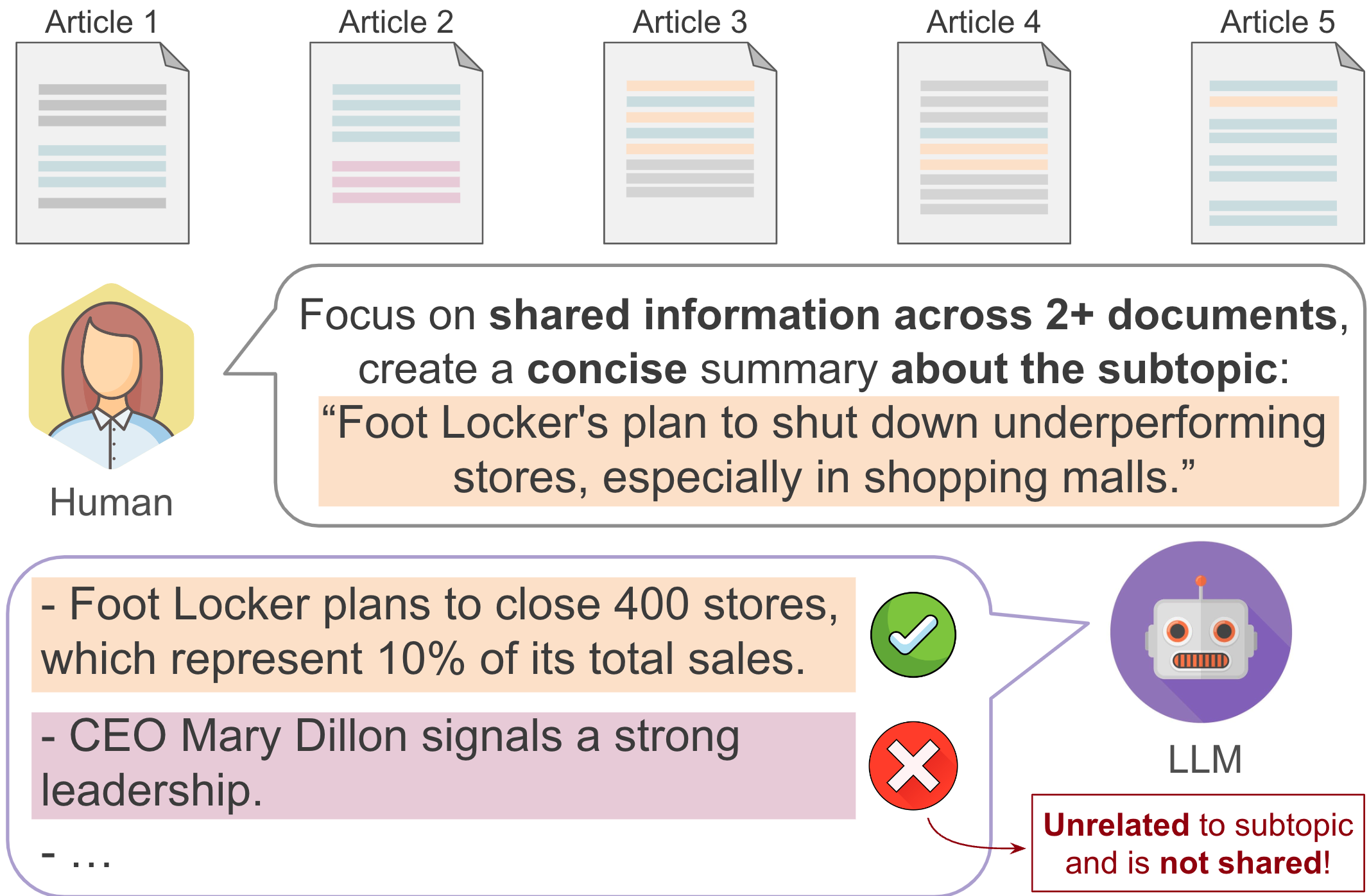}
    \caption{\textbf{An illustrative example of a summary generation from news articles}. Concerned about the credibility of the information, a human instructs the model to focus on shared, subtopic-related information. However, the \llm summarizes unrelated information that is not shared, raising concerns about the trustworthiness of \llms in \mds.}
    \label{fig:motivating-picture}
\postspace
\postspace
\end{figure}

However, despite the remarkable advances of \llms in various tasks~\citep{Bubeck2023SparksOA,zhao2024surveylargelanguagemodels}, 
the frequent generation of ungrounded yet plausible-sounding text, referred to as ``hallucinations'', undermines trust in their output~\citep{maynez-etal-2020-faithfulness,uluoglakci-temizel-2024-hypotermqa,Kalai-calibrated-lms-must-hallucinate2024}. 
%
%
While \llm \textit{hallucinations} have been extensively studied in single-document tasks~\citep{Ji-survey-of-hallucination-in-nlg-2023,huang2023surveyhallucinationlargelanguage}, resulting in the development of various evaluation benchmarks~\citep{lin-etal-2022-truthfulqa,wang-etal-2022-squality,yang-etal-2023-oasum} and taxonomies~\citep{rawte-etal-2023-troubling,zhang2023sirenssongaiocean,mishra2024finegrained,Dahl_2024}, little is known about how processing multiple documents affects the hallucinatory behavior of \llms in \mds.
%
%
Specifically, by focusing on the aggregated quality metrics of \llm outputs, including coverage~\citep{fabbri-etal-2019-multi,lu-etal-2020-multi-xscience,summhay--laban-et-al-2024} and faithfulness \citep{pu2023summarizationalmostdead,huang-etal-2024-embrace}, existing work offers limited insight into how the inherent challenges of \mds (\eg repeated and diversity of information) correlate with \llm hallucinations. 
Consider the motivating example in Figure \ref{fig:motivating-picture}: when summarizing the repeated information from multiple documents, the \llm fails to link overlapping information across documents and satisfy the subtopic condition (highlighted text in the figure). 

In this work, we comprehensively investigate the hallucinatory behavior of 5 popular \llms in \mds tasks. 
Focusing on news articles and dialogues summarization, we examine the properties of hallucinations produced by \llms when summarizing topic-specific information from multiple documents. 
Specifically, we aim to understand the patterns of hallucinations, their frequency, and their relationship with the number of input documents, as well as how these patterns change with the number of input documents and task focus (\eg summarizing undiscussed topics, or focusing the summary on repeated information). 
To this end, we design an evaluation protocol based on fine-grained annotations concerning relevant \textit{insights}---units of information---within each document. 
We use the insight-level annotations from the SummHay datasets~\citep{summhay--laban-et-al-2024}, designed for evaluating LLMs in long-context summarization, and create benchmarks with combinations of up to 10 documents in both the conversational and news domains.
Our goal here is to leverage the insight-level annotations to automatically assess the correctness of the \llm-generated summaries.\footnote{The code and data have been made publicly available: \url{https://github.com/megagonlabs/Hallucination_MDS}.}

Our empirical evaluation of 5 prominent \llms using the proposed evaluation protocol reveals that, up to 45\% and 75\% of the content in \llm-generated summaries is hallucinated in the news and conversation domain, respectively. 
We also observe that increasing the volume of input documents affects \llms differently: 
for instance, as the document count increases from 2 to 10, most models experience only marginal changes in hallucinated content ($\pm$ 5\%) whereas \fullgemini shows up to a 10\% increase. 
Examining the composition of \llm-generated summaries, we find that regardless of the number of input documents and summary focus, hallucinations are more likely to appear in the later sections of the summaries.
Furthermore, we find that both \shortgptfour and \shortchatgpt show a strong predisposition (about 44\% and 79.35\%, respectively) to generate summaries even when there is no topic-specific insight in the input, raising concerns about their tendency to fabricate content.

To better understand the characteristics of these hallucinations, we manually evaluate 700+ insights spanning all 5 evaluated models and discover that most errors stem from either failing to follow instructions (\eg topic-unrelated information and/or redundant information) or producing overly generic insights (\eg paraphrases of the topic). 
Based on these observations, we investigate the effectiveness of five simple post-processing approaches, including both rule-based and \llm-based approaches, in mitigating hallucinations.
We find these methods result in marginal improvements, on average: hallucinated content is decreased by up to 7\% points but at the cost of excluding relevant information by up to 6\%.
Together, these results suggest that reducing hallucinations without compromising model performance remains challenging, underscoring the need for further research to better understand and systematically prevent \llm hallucinations in \mds.

%
%
\section{Investigating Hallucination in \md}
\label{sec:proposed-eval}
Our goal is to shed light on the hallucination patterns in \llms when addressing multi-document summarization (\md). 
We begin by carefully defining the problem and its key assumptions. 
We then introduce two benchmarks, as well as the necessary evaluation metrics to discern between hallucinated and non-hallucinated \llm outputs. 

\subsection{Problem Formulation}
\label{ssec:methodology:problem-formulation}
We frame the \mds task as follows: 
given $N$ documents ($d_1, \dots, d_N$), and $K$ conditions ($c_1, ..., c_K$), an \llm must generate a summary $\hat{\mathbf{y}}$ such that it satisfies all $K$ conditions and is grounded in the documents. 
Examples of conditions include matching specific topics, adhering to length constraints, or following particular writing styles.
For simplicity, we assume that the $N$ documents in our formulation are of \textit{the same type} (\eg all documents are news articles), rather than from diverse sources (\eg job descriptions and resumes, doctor notes and prescriptions). 
The summarization of multiple documents may involve capturing information that is either diverse~\citep{huang-etal-2024-embrace}, contradictory, or common across documents.
The latter is especially important for combating misinformation, as it emphasizes ``trustworthy'' information supported by multiple sources. 
To analyze model behavior in both contexts, we require \textit{fine-grained annotations} of the \textit{information units} contained in each document. 

\subsection{Dataset Creation}
\label{ssec:methodology:dataset-creation}

Recently, \citet{summhay--laban-et-al-2024} introduced two \md datasets with \textit{insight-level annotations}.\footnote{The authors define \textit{insights} as ``units of information''.}
The datasets cover two distinct domains---\textit{news} and \textit{conversation}---each containing 500 documents, evenly distributed across 5 topics, with documents averaging around 750 words. 
In particular, the news dataset (\newsdataset) is entity-centric and quantitative, featuring brands (\eg Tesla), banks (\eg JP Morgan), celebrities (\eg Elon Musk), along with numbers and dates. 
In contrast, the conversation dataset (\convdataset) involves everyday scenarios with 2 to 4 participants, such as medical appointments and debates. 
Originally developed to evaluate \llms in long-context tasks (\eg retriever-augmented generation, multi-document summarization), the documents were designed to ensure insights are repeated across at least 6 per topic and categorized into various subtopics, with no contradictory insights present.

By selecting \newsdataset and \convdataset as testbeds, we can exploit their insight-level annotations to systematically assess model behavior across different settings, such as number of input documents and task focus. 
To control for input length~\citep{li2024longcontextllmsstrugglelong} while evaluating model behavior, we organize documents in the SummHay dataset into sets of $N$ documents, referred to as \textit{combinations of $N$} or \textit{$N$-documents} (see Figure \ref{fig:evaluation-methodology} in Appendix for an illustration of our benchmark creation process).
Specifically, given a corpus with insight-level annotations and a subtopic $q$, we create our benchmarks by grouping documents into combinations of $N$ where multiple subtopic-related insights co-occur in two or more documents.
Because we have the insight-level annotations, we can automatically identify the ground truth insights related to subtopic $q$ (called \textit{reference insights}). 
To fully benefit from these fine-grained annotations, we instruct \llms to succinctly\footnote{
Unlike \citeauthor{summhay--laban-et-al-2024}, we do not instruct \llms to limit their generation to a fixed number of bullet-points. 
Instead, we study their behavior ``in the wild'' when the exact number of insights is unknown. 
For further details related to prompts, refer to Appendix \ref{app:sec:prompt-selection}.
} summarize the information as individual insights, presented in the form of bullet-point lists.
We coin this setting \subtopic. 
To further investigate how the model handles common information, we introduce a \subtopictrust setting by refining the prompt and restricting the reference insights to only those that are shared across documents.

With the intent of investigating the impact of document volume in the model's tendency to hallucinate, we employ the previous methodology to generate combinations of N for both \newsdataset and \convdataset across five different combination sizes $N=\{2, 3, 4, 5, 10\}$. 
Since analyzing model behavior across all existing combinations-subtopic pairs is prohibitively expensive, we conduct our analysis on 500 randomly selected combinations for each $N$. 
Refer to Appendix \ref{app:sec:dataset} for additional information on the resulting benchmarks.

Note that while the proposed methodology applies to any \mds dataset with high-quality insight-level annotations, such information is rarely available.
Therefore, we limit our evaluation to the SummHay dataset and leave the evaluation of other datasets, including non-English datasets and those with contradictory insights, for future work.

\subsection{Automatic Evaluation}
\label{ssec:methodology:evaluation}
With reference insights for each combination, we can automatically evaluate the \textit{correctness} of \llm outputs, distinguishing between hallucinatory and non-hallucinatory content. 
The correctness metric should assess whether predicted insights capture all essential information from the reference insights while avoiding the inclusion of irrelevant details~\citep{huang-etal-2024-embrace}. 
This ensures that predicted insights do not contain unsupported or potentially fabricated information. 

Previous work uses an \textit{\llm-as-a-judge} approach~\citep{zheng2023judgingllmasajudgemtbenchchatbot} to assess whether reference insights are fully, partially, or not covered by the predicted insights, showing strong correlation with human evaluations~\citep{summhay--laban-et-al-2024}. 
However, this metric only measures reference coverage and ignores the validity of additional details in predictions. 
To enhance this, we assess how well predicted insights align with reference insights by applying the original metric twice with swapped inputs---once for reference insight and once for predicted insight—--yielding two coverage labels per pair. 
We then combine these into a single correctness label using a conservative approach based on the order: \nocov $\prec$ \partcov $\prec$ \fullcov. 
By selecting the worst label (i.e., select \nocov over \partcov and select \partcov over \fullcov), we ensure the final assessment captures discrepancies related to missing relevant information or any potentially unfaithful content that is generated by the \llm.
%
%
\begin{figure*}[tb!]
    \centering
    \begin{subfigure}[b]{0.5\columnwidth}
        \centering
        \includegraphics[width=\columnwidth]{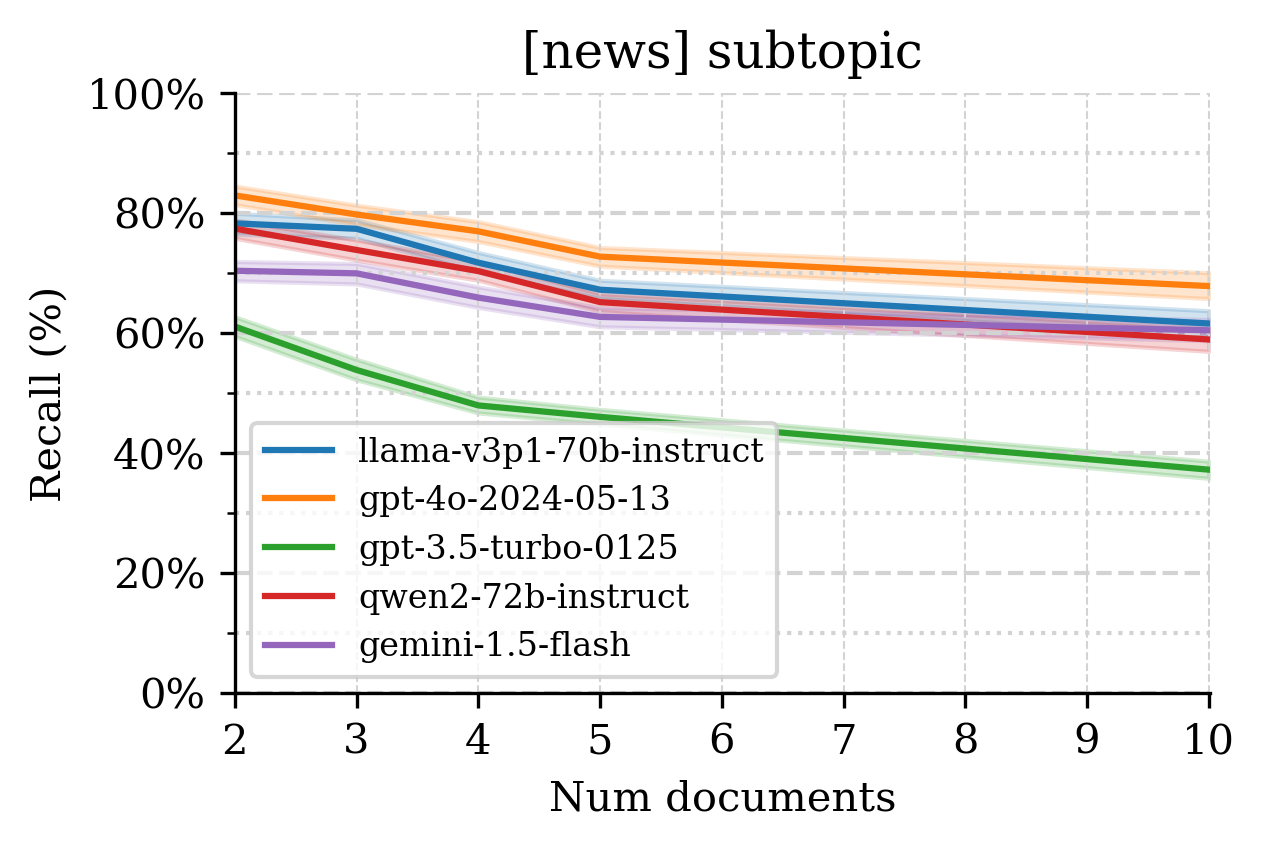}
        \caption{Recall {(news)}}
        \label{fig:main:subtopic:recall:news}
    \end{subfigure}
    \begin{subfigure}[b]{0.5\columnwidth}
        \centering
         \includegraphics[width=\columnwidth]{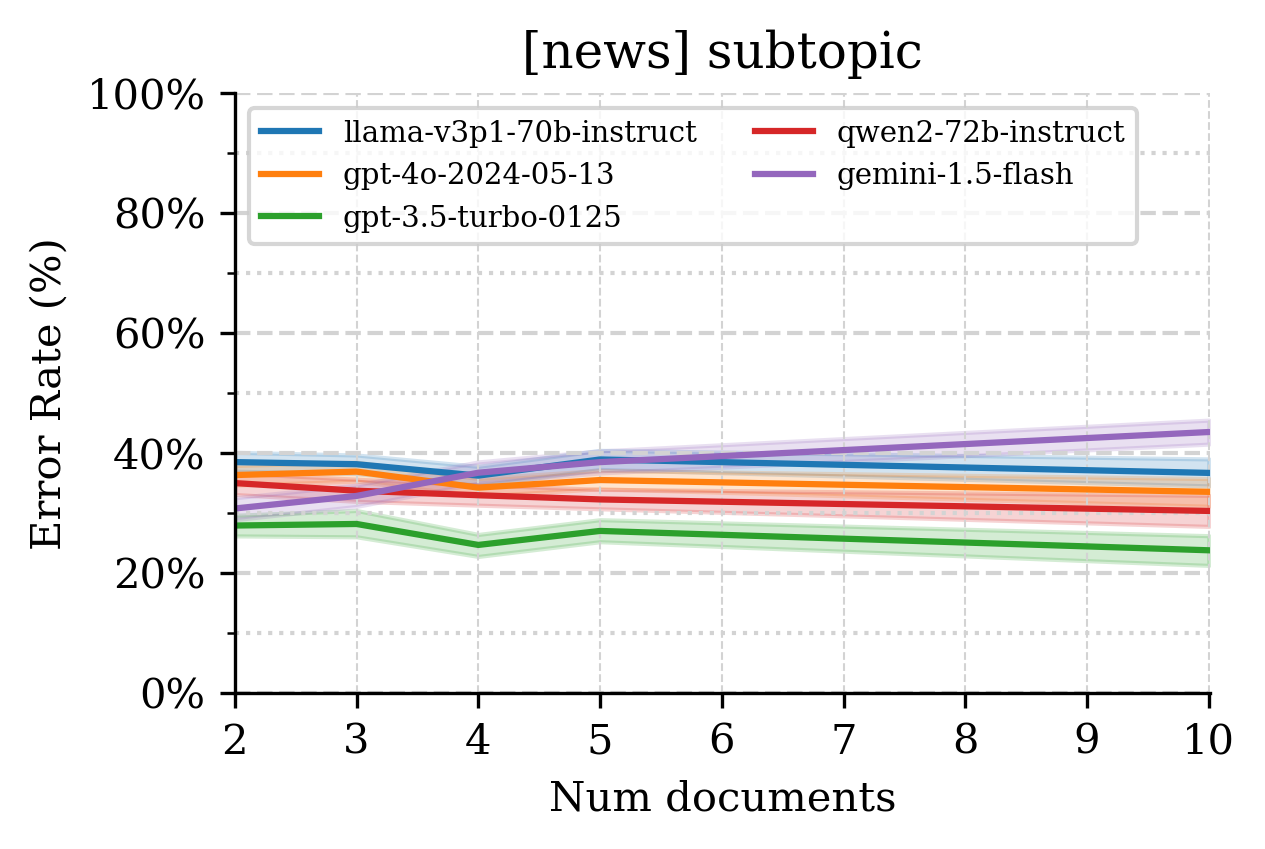}
        \caption{Error Rate  {(news)}}
        \label{fig:main:subtopic:err-rate:news}
    \end{subfigure}
    \begin{subfigure}[b]{0.5\columnwidth}
        \centering
         \includegraphics[width=\columnwidth]{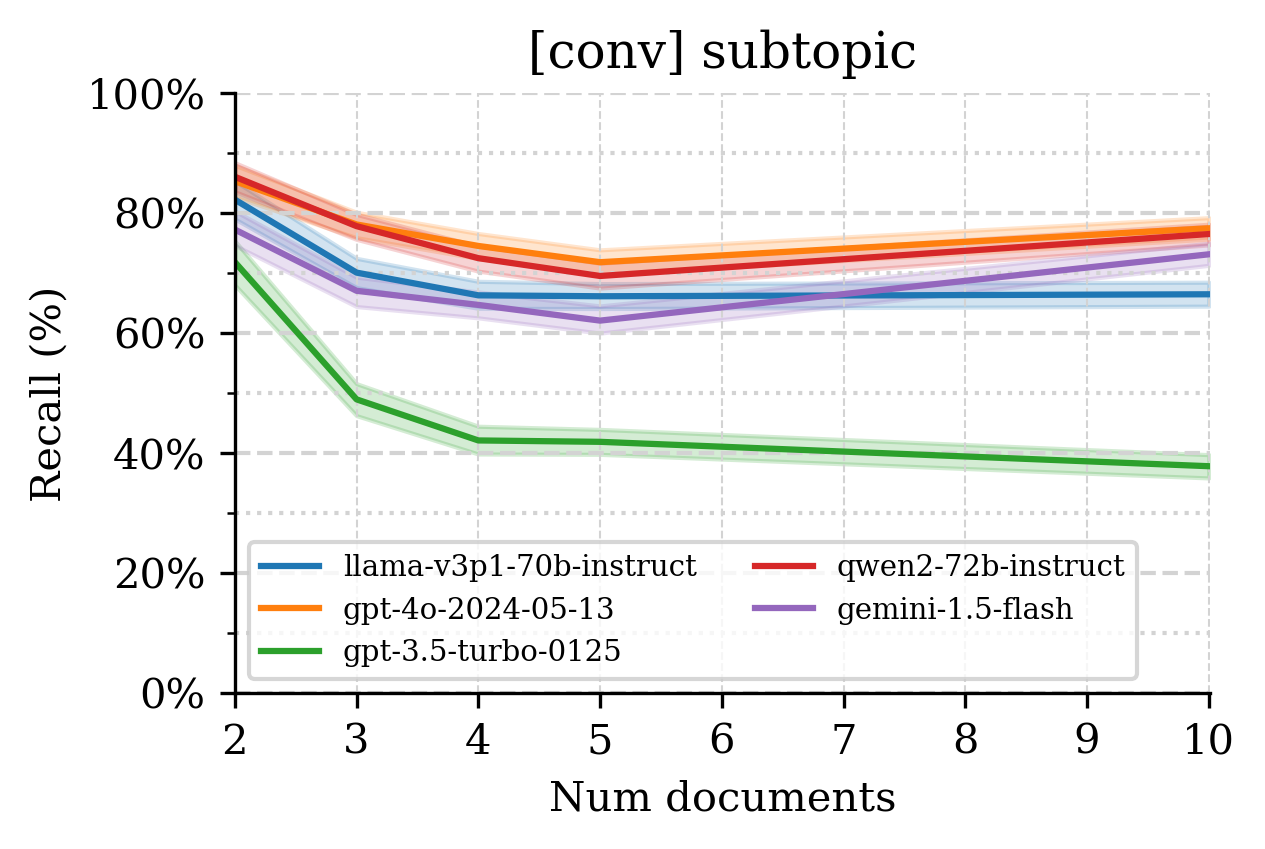}
        \caption{Recall {(conv)}}
        \label{fig:main:subtopic:recall:conv}
    \end{subfigure}
    \begin{subfigure}[b]{0.5\columnwidth}
        \centering
         \includegraphics[width=\columnwidth]{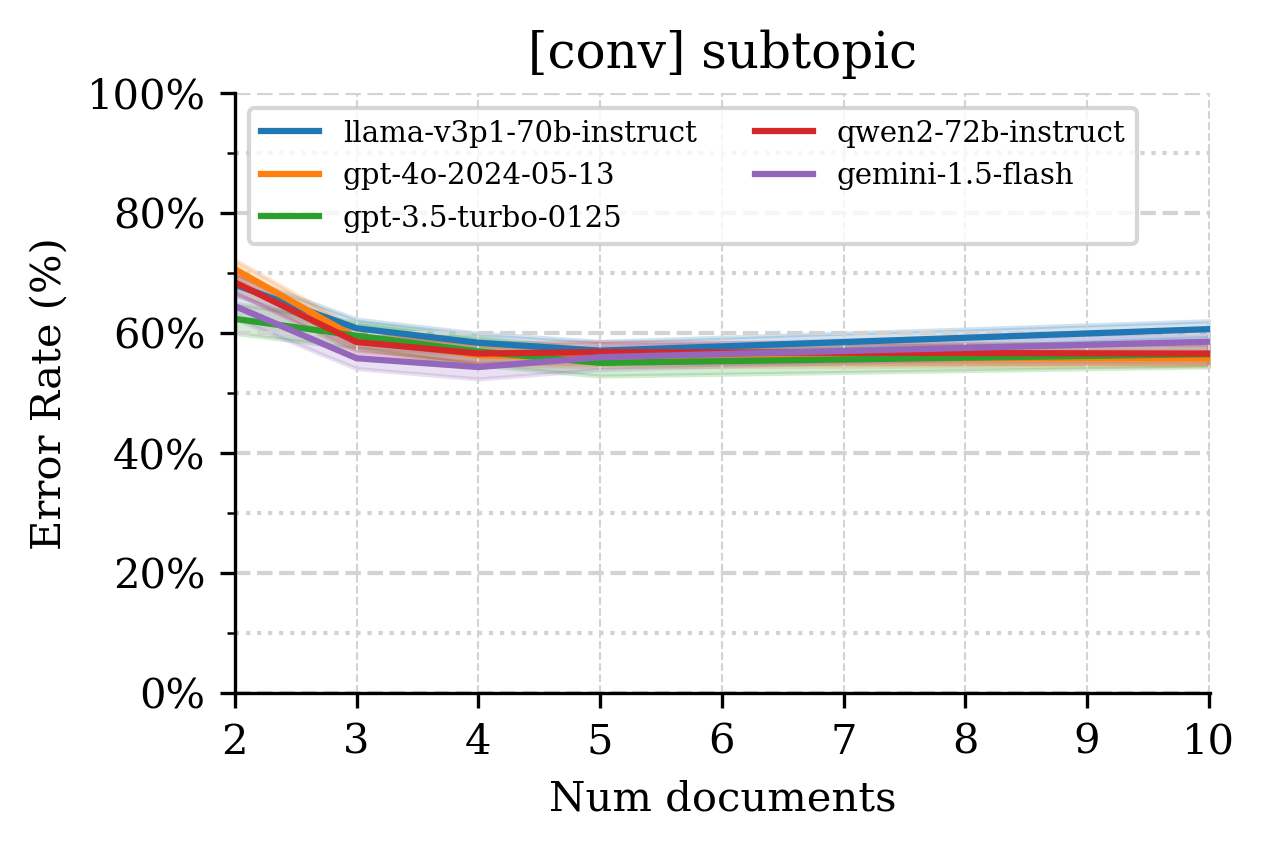}
        \caption{Error Rate {(conv)}}
        \label{fig:main:subtopic:err-rate:conv}
    \end{subfigure}
    \caption{\textbf{Performance metrics as a function of input documents counts in the \subtopic setting}. Each line represents the mean value, with shaded areas indicating the 95\% confidence intervals. Generally, recall drops significantly as document count increases, while average error rate changes only slightly across models and domains.}
    \label{fig:main:subtopic:metrics}
\postspace
\minipostspace
\end{figure*}

\subsection{Metrics}
\label{ssec:methodology:metrics}
A key question in \mds tasks is whether \llm-generated summaries include all the correct information. 
To estimate average \llm correctness, we calculate \textit{macro-recall} by determining the fraction of reference insights covered in each generated summary and averaging these scores across all summaries to obtain a single score. 
We also report the average fraction of hallucinated content in generated summaries by computing the average proportion of predicted insights that do not correspond to any reference insight across all summaries. 
This metric is referred to as the false discovery rate (\textit{macro-FDR}).
Both metrics range from 0 to 1, with a perfect \llm scoring 1 for macro-recall and 0 for macro-FDR. 
For simplicity, we refer to \textit{macro-recall} as \textbf{recall} and \textit{macro-FDR} as \textbf{hallucination rate}.

%
%
\section{Experimental Settings}
\label{sec:experiment_setup}

Before delving into the details of our investigation into \llm behavior in \mds, we briefly discuss key aspects of the conducted evaluation:  

\textbf{Models. }
We examine the capabilities of 5 popular \llms, that span both open-source and closed-source models.
As part of the open-source models, we evaluate the instruction-tuned version of \shortllama~\citep{llama-3.1-blogplost} and \shortqwen~\citep{yang2024qwen2technicalreport} models due to their competitive performance and instruction-following capabilities \citep{chiang2024chatbotarenaopenplatform,open-llm-leaderboard-v2,llm-hall-index-rag-2024}. 
As for the closed-source models, we assess OpenAI's~\fullchatgpt and \fullgptfour models~\citep{openai-gpt-4o-blogpost}, as well as Google's \fullgemini~\citep{geminiteam2024gemini}.
More details are provided in Appendix \ref{app:sec:model-access-details}.  

\textbf{Automatic Evaluation. }
As previously mentioned, we adopt a few-shot \textit{\llm-as-a-judge} approach. 
However, instead of using \texttt{gpt-4o} like \citet{summhay--laban-et-al-2024}, we resort to a competitive yet more budget-friendly option---\texttt{gpt-4o-mini-2024-07-18}. 
The authors manually annotated 100 insight pairs and found strong alignment between both models, confirming the suitability of \texttt{gpt-4o-mini-2024-07-18} as an evaluator. 
Refer to Appendix \ref{app:sec:automatic-metric-validation} for more details. 

\textbf{Metrics. }
There is a mismatch between metrics: recall and hallucination use binary labels (correct/incorrect), but the \llm-based metric uses three coverage labels (\nocov, \partcov, \fullcov). 
To report recall and hallucination, we must map these three labels into correctness labels. 
We find the exact mapping has minimal impact (<5\%) in the news domain but a larger effect ($\sim$30\%) in the conversation domain. 
In the following sections, we opt for an optimistic approach, treating both partially and fully covered insights as correct (more details in Appendix \ref{app:ssec:impact-cov-labels}).

%
%
\section{Experimental Results}
\label{sec:experimental-results:investigation-results}

In this section, we first investigate the extent of \llms hallucinations in \mds under the proposed \subtopic and \subtopictrust settings. 
Then, we analyze \llms' hallucinatory behavior when synthesizing information on non-existent topics and investigate the link between hallucinated content and characteristics of both the input and output (we use the prompts listed in Appendix \ref{app:sec:prompt-selection}).
\begin{figure*}[tb]
    \begin{subfigure}[b]{0.5\columnwidth}
        \centering
        \includegraphics[width=\columnwidth]{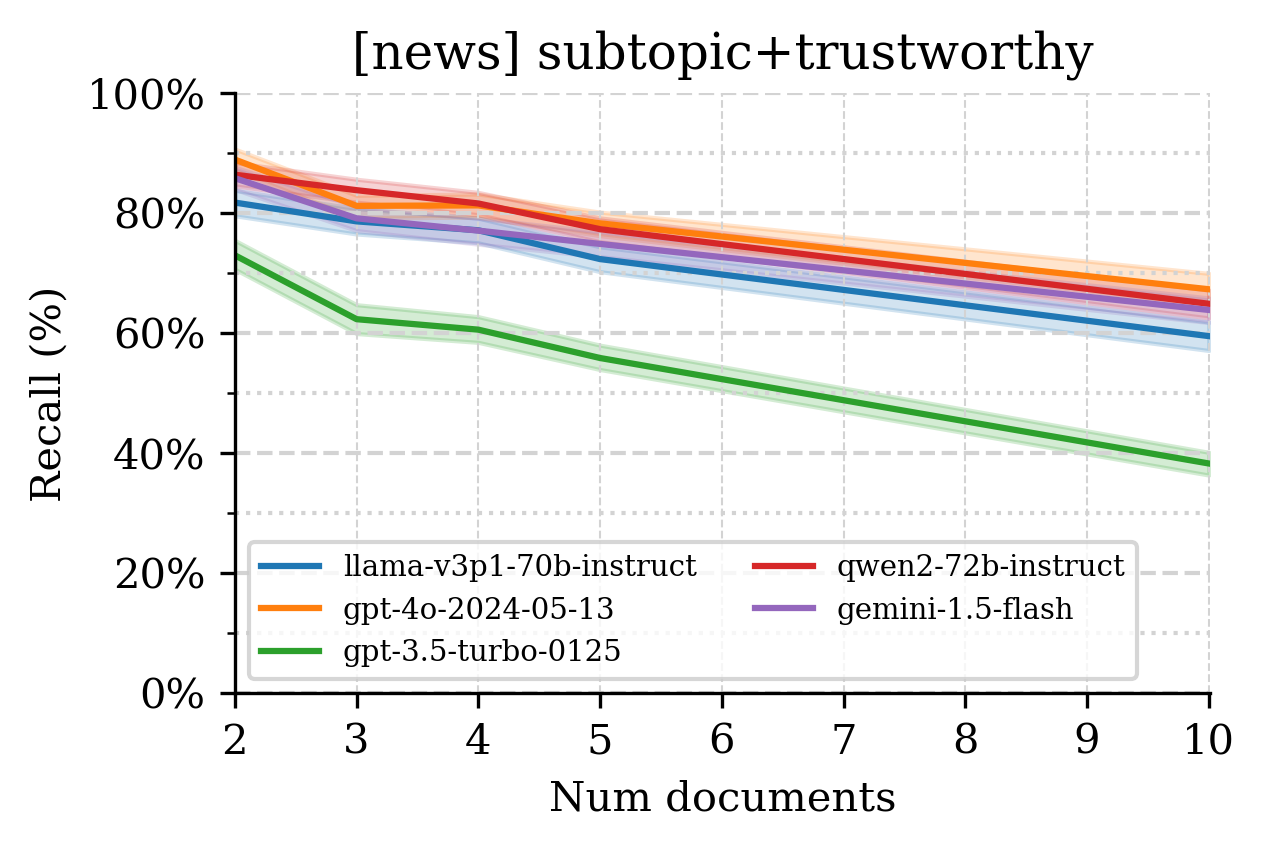}
        \caption{Recall {(news)}}
        \label{fig:main:subtopic+shared:recall:news}
    \end{subfigure}
    \begin{subfigure}[b]{0.5\columnwidth}
        \centering
         \includegraphics[width=\columnwidth]{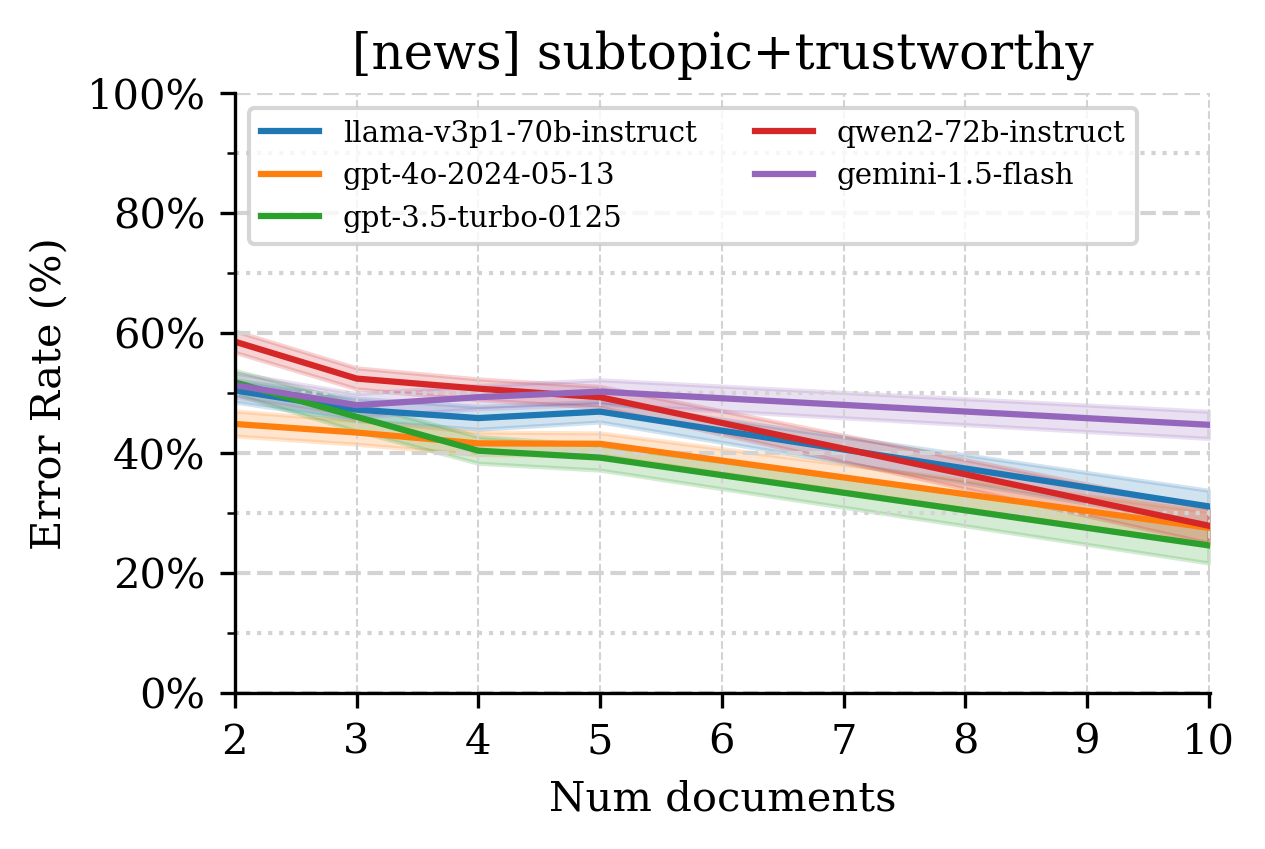}
        \caption{Error Rate {(news)}}
        \label{fig:main:subtopic+shared:err-rate:news}
    \end{subfigure}
    \begin{subfigure}[b]{0.5\columnwidth}
        \centering
        \includegraphics[width=\columnwidth]{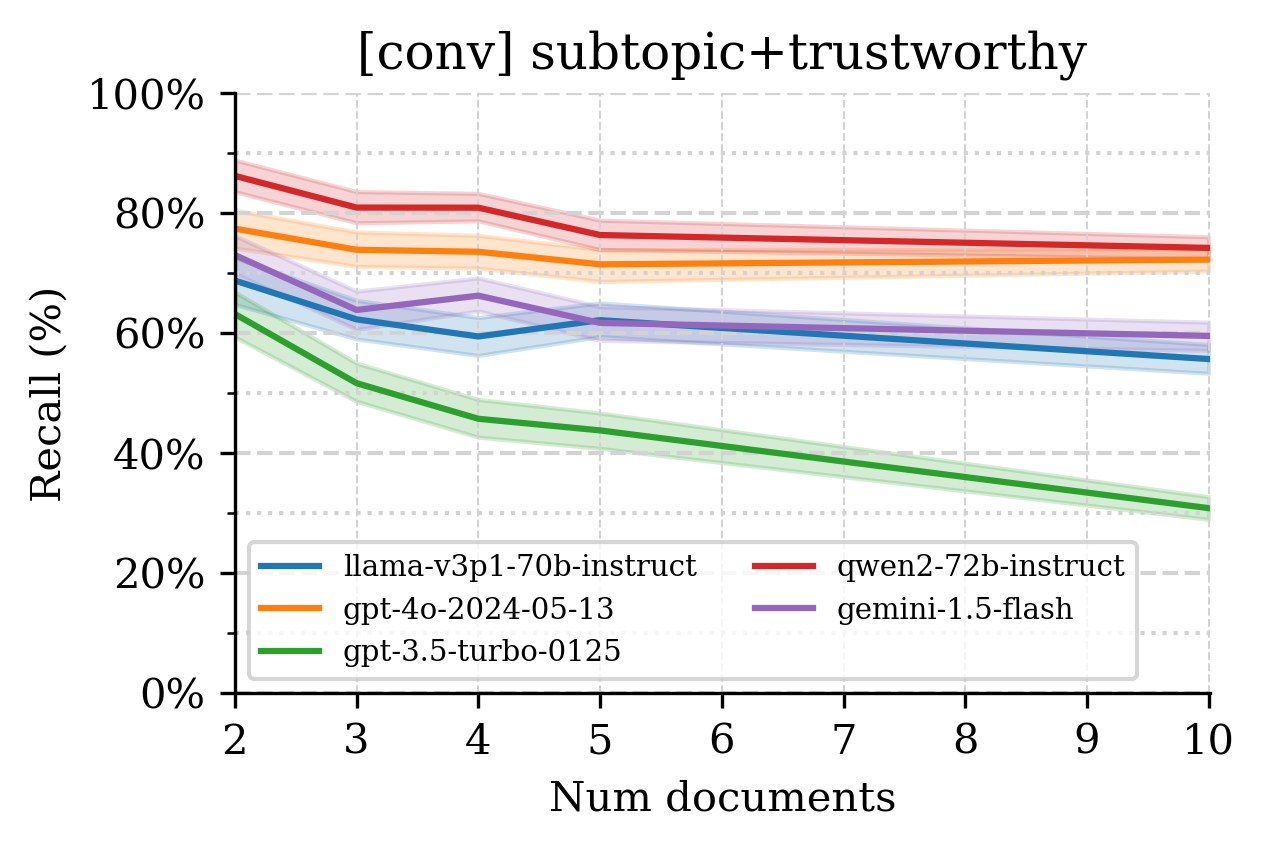}
        \caption{Recall {(conv)}}
        \label{fig:main:subtopic+shared:recall:conv}
    \end{subfigure}
    \begin{subfigure}[b]{0.5\columnwidth}
        \centering
         \includegraphics[width=\columnwidth]{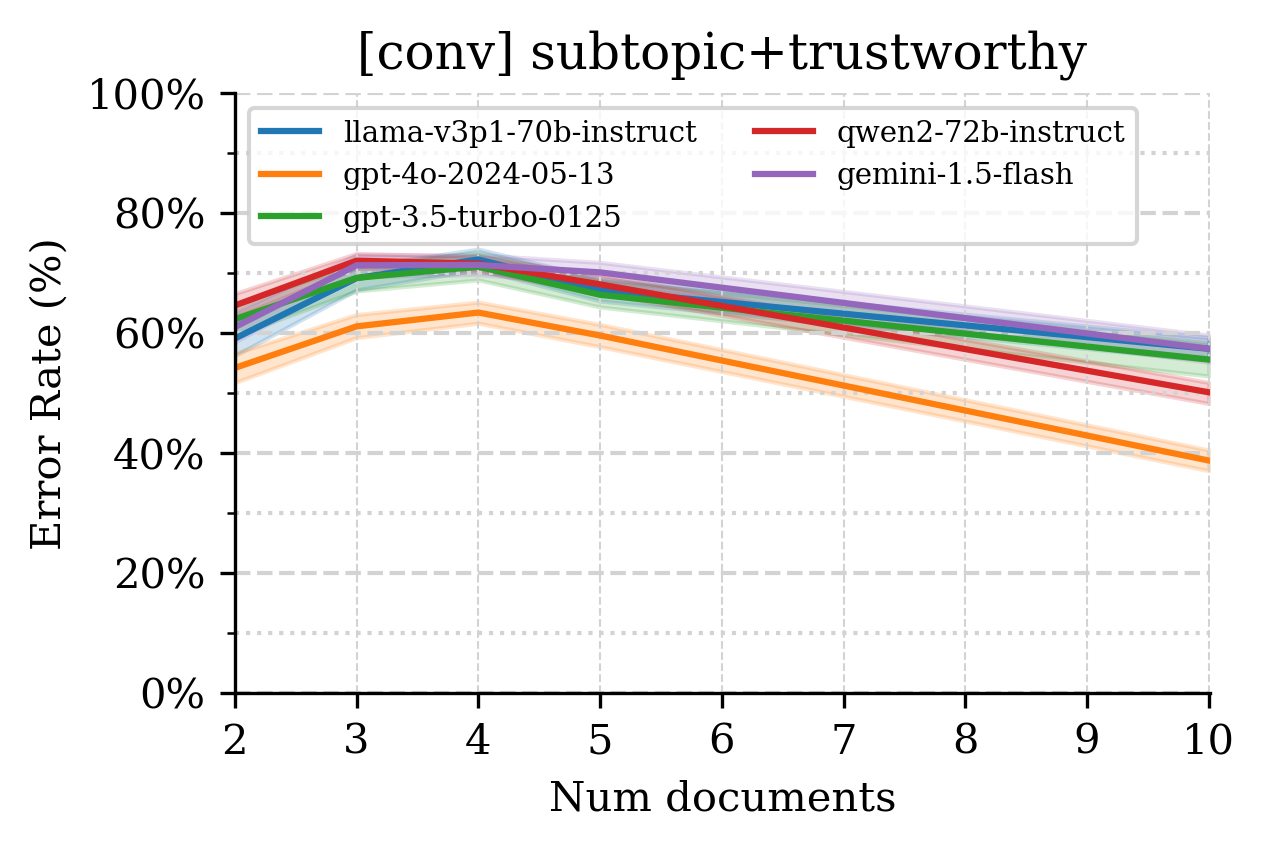}
        \caption{Error Rate {(conv)}}
        \label{fig:main:subtopic+shared:err-rate:conv}
    \end{subfigure}
    \vspace{-0.5em}
    \caption{\textbf{Performance metrics as a function of the number of input documents in the \subtopictrust setting}. Each line represents the mean value, with shaded areas indicating the 95\% confidence intervals.}
    \label{fig:main:subtopic+shared}
\postspace
\minipostspace
\end{figure*}

\subsection{\llms hallucinate in \md}
\label{ssec:investigate:subtopic}
To investigate \llms hallucinatory behavior in \md, we instruct each \llm using the \subtopic prompt to generate a summary for 500 examples of the datasets created in Section \ref{ssec:methodology:metrics}.
We then evaluate the recall and hallucination rate associated with each summary. 
Our main findings are as follows:\footnote{Refer to Appendix \ref{app:sec:additional-results} for additional results.}

\textbf{\textit{\llms exhibit substantial hallucination rates}} (see Figure \ref{fig:main:subtopic:metrics}). 
Across models and combination size ($N$), we observe an average hallucination rate greater or equal than $20$\% and $52$\% (and up to 45\% and 75\%) for the news and conversation domains, indicating that a non-negligible portion of \llm-generated content is hallucinated. 
%
%
Moreover, we observe that, \textbf{\textit{hallucination rate changes only marginally with input size.}}
Intuitively, increasing the number of documents introduces distracting information, which can affect the \llm-generated summaries---potentially, resulting in lower recall and higher error rates. 
Figure \ref{fig:main:subtopic:metrics} shows that, although there is an overall downward trend in recall, the error rate remains almost constant ($\pm 5$\%), increasing only for \shortgemini ($<$10\%). 
Since we withhold the expected number of insights from \llms, we hypothesize that the observed patterns stem from a mismatch between the number of predicted insights and reference insights.
Upon further analysis (see Appendix \ref{app:ssec:number-insights}), we validate this pattern between observed recall drops and the ratio of predicted-to-reference insights. 

\textbf{\textit{Models make more mistakes when summarizing conversations than news articles.}}
Overall, we find a 20-30\% hallucination rate disparity between the two domains. One possible explanation may be rooted in the dataset properties. 
In particular, the \newsdataset is entity-centric, discussing various concerns about common celebrities and companies (\eg Twitter), whereas the \convdataset is more contextual discussing multi-turn everyday interactions between different participants. 
We indeed verify this is the case in practice, after manually inspecting 25 insights from each domain. 

\subsection{Shared Insights, Greater Errors}
\label{ssec:investigate:subtopic-shared}

Compared to the previous \subtopic setting, analyzing model behavior in \subtopictrust requires adjusting the evaluation process and summarization prompt to focus only on shared insights. 
As before, we prompt models to summarize all 500 examples per benchmark and measure recall and hallucination rates. Our findings are as follows:

\textbf{\textit{\llms generate shorter summaries but hallucinate more in general.}}
To understand how responsive models are to the \textit{shared instruction}, we compare summary lengths between \subtopic and \subtopictrust settings (see Appendix \ref{app:ssec:number-insights}). 
Summaries in \subtopictrust are shorter, suggesting models respond to the \textit{shared} instruction. 
However, shorter summaries do not guarantee quality: hallucination rates are, on average, higher in \subtopictrust than \subtopic (news: +10.47\%, conv: +4.20\%), indicating models may struggle to identify shared insights.
We also observe that, \textbf{\textit{\llms struggle to identify subtopic-related shared insights.}}
Compared to the \subtopic scenario, we observe an average increase in summarization performance in news domain (+6.93\%) but a slight decrease in the conversation (-2.91\%). %
Overall recall trends remain the same: larger combination sizes lead to up to a 33\% drop, with the sharpest declines in \shortchatgpt. 
Notably, \shortqwen is mostly on par with, sometimes superior to, \shortgptfour (58.9-86.5\% \textit{vs} 67.3-86.9\%). 
However, unlike \shortgptfour, \shortqwen generates longer summaries and has a higher hallucination rate (see Appendix \ref{app:ssec:number-insights}).
\subsection{Summarizing the Unsummarizable}
\label{ssec:investigate:adversarial}

Thus far, we have examined model behavior in well-defined scenarios where, by design, models summarize information that is known to exist in the input documents. 
However, this assumption may not hold true in practice. 
For instance, when summarizing opinions~\citep{angelidis-etal-2021-extractive,amplayo-etal-2021-aspect,hosking-etal-2023-attributable}, it is possible that models are instructed to synthesize information along specific aspects (\eg ``product reliability'' or ``location of the resort'') that were not mentioned. 

Ideally, a reliable \llm would avoid generating summaries in such \textit{adversarial} cases, but it remains uncertain whether the evaluated \llms will be able to do so, especially given their sycophantic tendencies~\citep{sharma2024towards,rrv-etal-2024-chaos}. 
To assess \llms' ability to abstain from generating incorrect information, we modify 250 samples from the proposed benchmarks and pair them with a subtopic $q$ that, while still related to the theme, it is not explicitly discussed in the input documents. We instruct \llms to output ``No insights found'' when relevant insights are absent and measure how often they \textit{abstain} from summary generation.

\begin{figure}[tb]
    \begin{subfigure}[b]{\columnwidth}
        \centering
        \includegraphics[width=0.95\columnwidth]{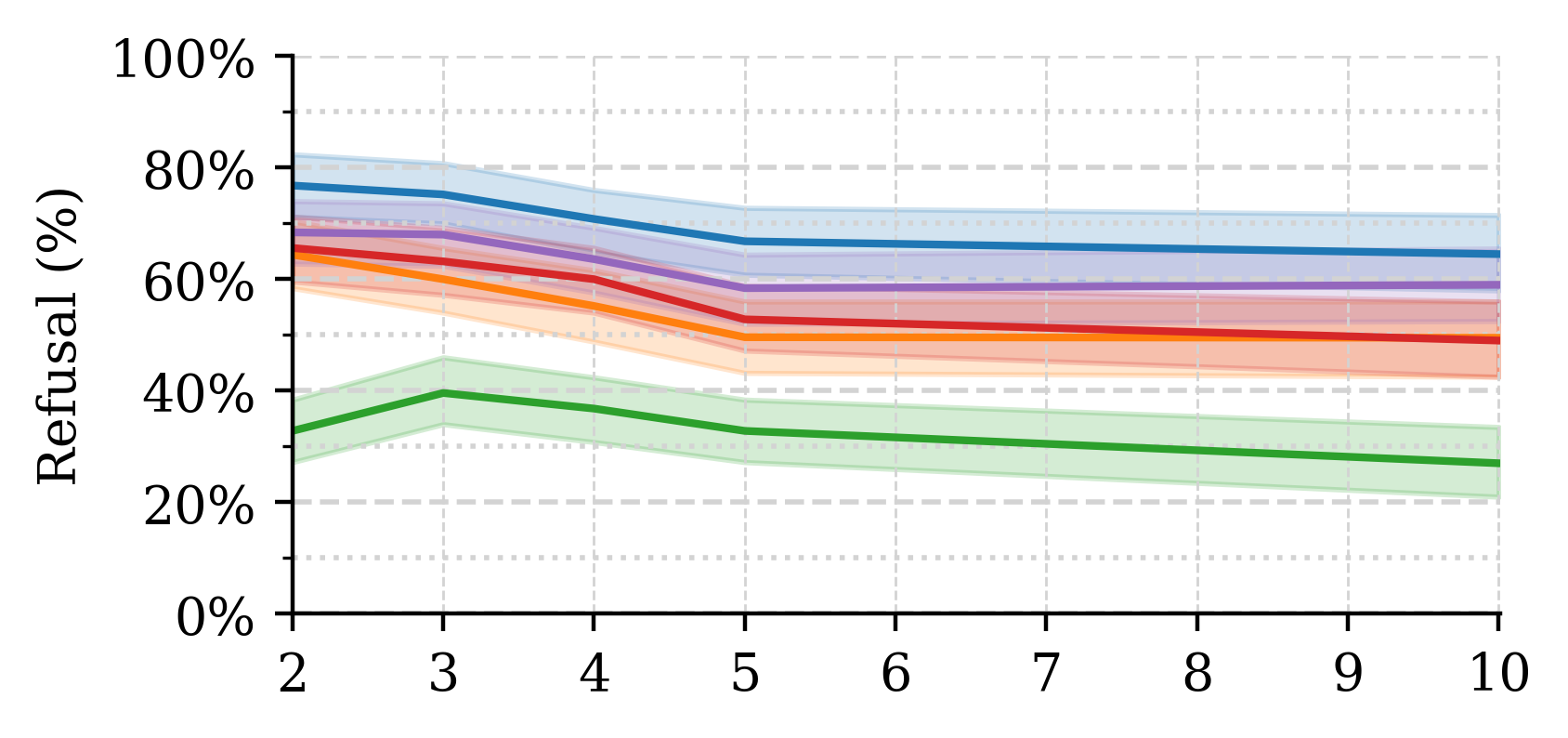}
        \label{fig:main:adversarial:news}
    \end{subfigure}\\[-1.4em]
    \begin{subfigure}[b]{\columnwidth}
        \centering
        \includegraphics[width=0.95\columnwidth]{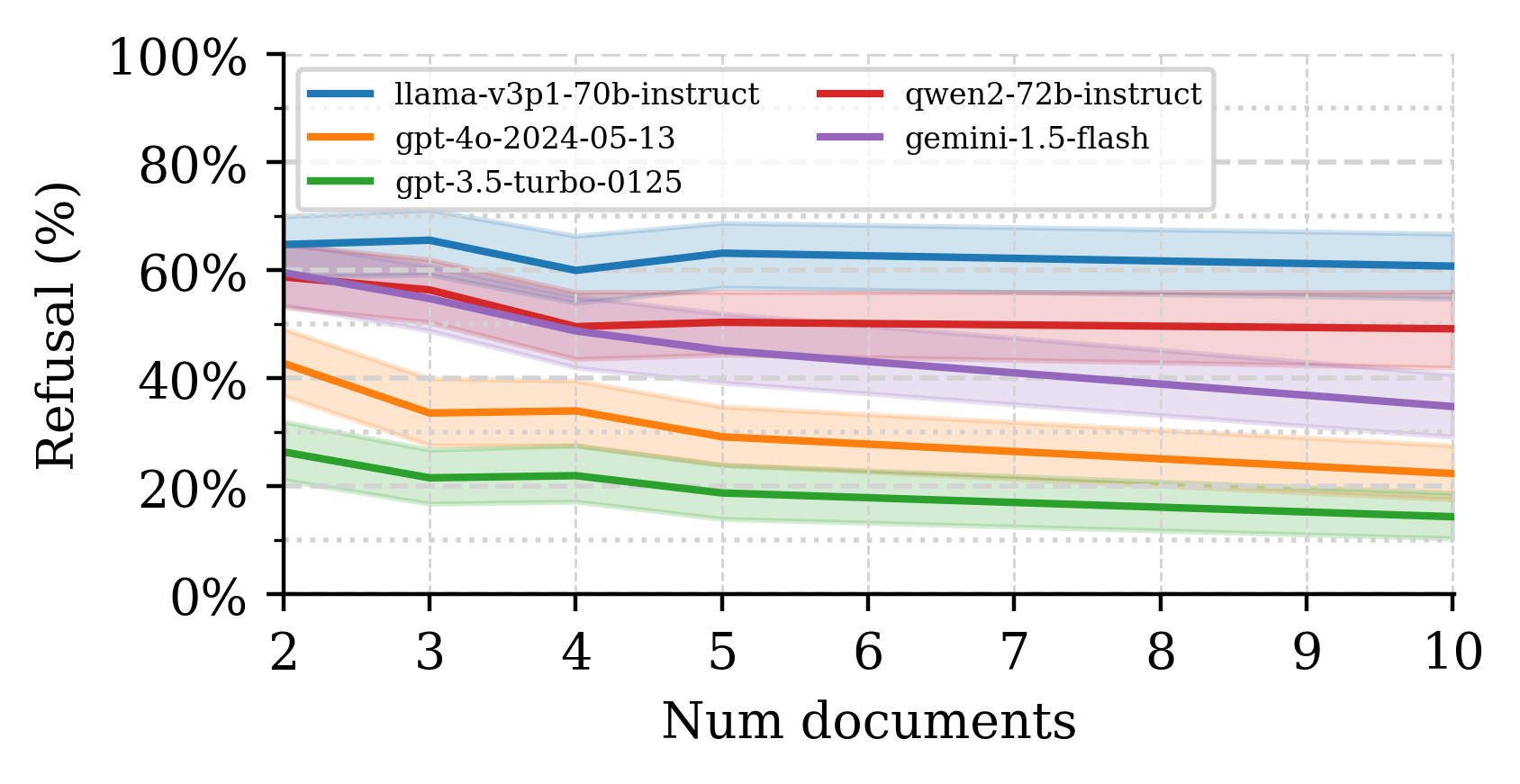}
        \label{fig:main:adversarial:conv}
    \end{subfigure}
    \vspace{-2em}
    \caption{\textbf{Mean and $95$\% confidence intervals of summary refusal rate (\%) for the news (top) and conversation (bottom) domains}. Notably, OpenAI models perform the worst, while \shortllama consistently abstains from generating summaries over 60+\% of the time, regardless of document count.}
    \label{fig:main:adversarial}
\postspace
\minipostspace
\end{figure}

\textbf{\textit{Models mistakenly generate summaries even if no relevant information is provided.}} 
Figure \ref{fig:main:adversarial} shows that models actually generate summaries for non-existent subtopics: \llms only abstain in 20.65\% (\shortchatgpt, conversation) and up to 71.08\% (\shortllama, news).
Moreover, \textbf{\textit{as the document count increases, \llms are more prone to mistakenly generate summaries.}}
In general, our findings show a sharp decline in \llms' capability to output ``No insights found'' as the number of input documents increases---this decline is especially pronounced in the conversation setting.
Finally, we observe that open-source models tend to make \textit{fewer mistakes} compared to proprietary models. 
Notably, \shortllama outperforms other models (71.08\% on average) and is least affected by input size (<8\% drop). 
In contrast, OpenAI models show lower performance and greater sensitivity to the number of documents, incorrectly generating summaries in up to 44\% and 79.35\%.
%

\subsection{Which documents do errors stem from?}
\label{ssec:investigate:error-vs-input}

So far, we have presented empirical evidence that models consistently hallucinate, irrespective of the number of input documents or task focus.  
In this section, we investigate the source of model hallucinations by examining how they correlate with the input documents, \ie we ask the question \textit{where do hallucinations come from}?\footnote{Prior work has studied this question with respect the faithfulness score~\citep{huang-etal-2024-embrace}.}
We tackle this question by first determining which input document (if any) the hallucinated insight is copied from. 
In particular, we modify the string matching approach used to determine data contamination in the GPT-4 paper~\citep{openai-gpt-4o-blogpost}, leaving the exploration of other techniques~\citep{xu2024benchmarkdatacontaminationlarge} for future work.
Specifically, we assume that an hallucinated insight $i$ originates from a document $d$ if any 50-character string from $i$ matches a string in $d$. 

\begin{figure}[tb]
    \centering
     \includegraphics[width=\columnwidth]{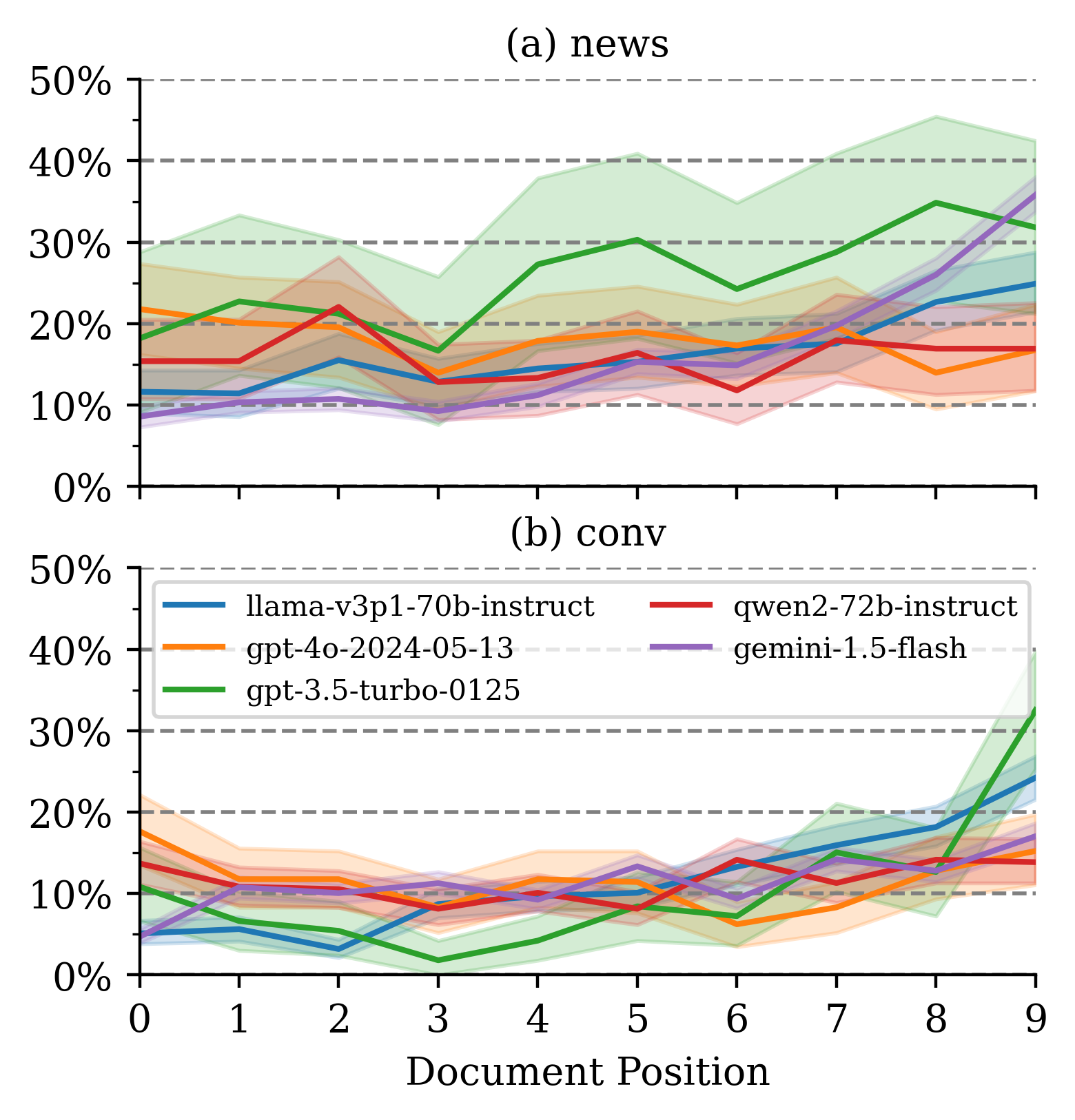}
    \caption{\textbf{Likelihood of an insight coming from a document (y-axis) based on its position in the input (when summarizing 10 documents)}. \shortchatgpt, \shortllama and \shortgemini mistakes seem to be more likely to originate from later documents on average than from earlier ones.}
    \label{fig:main:investigate:error-vs-input:ndocs-10}
\postspace
\minipostspace
\end{figure}

We find that \textbf{\textit{3 out of 5 models exhibit a slight recency bias towards the last documents}}. 
As observed in Figure \ref{fig:main:investigate:error-vs-input:ndocs-10}, this recency bias is particularly prominent in \shortchatgpt and \shortgemini in the news domain, but also present in \shortchatgpt and \shortllama in the conversation domain. 
We hypothesize that such biases may emerge from the proximity to the list of instructions or due to the models' positional encoding, whose exploration we leave for future work.

\subsection{Output order: A Proxy for Correctness?}
\label{ssec:investigate:error-vs-output}

Having explored how errors relate to the input, we now turn our attention to how hallucinations manifest in \llms outputs. 
Given that models are instructed to produce summaries in bullet-point format, we can gauge the likelihood of the i-th bullet-point (\ie predicted insight) being hallucinated. 
To be more precise, we assess the relationship between the insights' position and their accuracy rate. 

\begin{figure}[tb]
\centering
\includegraphics[width=0.95\columnwidth]{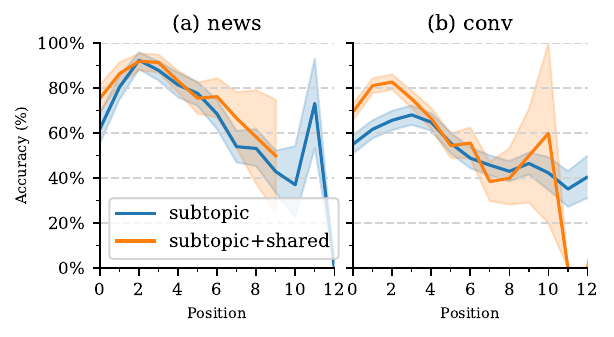}
\caption{\textbf{Accuracy rate of \shortgptfour generated insights by position (when summarizing 10 input documents)}. Each solid line shows the mean, with shaded areas representing 95\% confidence intervals. Overall, accuracy rate declines as insight position increases.}
\label{fig:main:investigate:error-vs-output:ndocs-10}
\postspace
\minipostspace
\end{figure}

We observe that \textbf{\textit{insights positioned earlier in the summary are more likely to be accurate than those located later}} (see Figure \ref{fig:main:investigate:error-vs-output:ndocs-10}), consistent with previous findings~\citep{min-etal-2023-factscore,chen2023understandingretrievalaugmentationlongform}.
In Appendix \ref{app:sec:data-output}, we show that this pattern holds across all summarization models, regardless of input document count. 
Interestingly, the opposite pattern appears in the top three summary positions. 
Manual inspection reveals that lower accuracy at the start is due to the generation of broader scope insights (\eg opening remarks or preliminary statements).
We also find that \shortgptfour and \shortgemini tend to generate 1 or 2 ``takeaway'' or ``concluding'' insights at the end of the summary to emphasize the main point or lesson to remember. 
We hypothesize this structure may stem from discourse coherence~\citep{Jurafsky-discourse-coherence,zhu-etal-2024-couda} or the models' ability to mimic input document patterns. Future research could explore these possibilities further.

\section{What type of mistakes do \llms make?}
\label{sec:error-taxonomy}
\begin{table*}[ht!]
    \small
    \centering
    \caption{\textbf{Definition of the hallucination types found in \llm-generated summaries across both domains}.}
    \label{tab:hal-tax}
    \vspace{-0.8em}
    \begin{tabular}{p{0.15\textwidth} p{0.80\textwidth}}
        \toprule
        \multicolumn{1}{c}{\textbf{Type}} & \multicolumn{1}{c}{\textbf{Definition}}\\
        \midrule
        \sloppy\raggedright Pedantic & Insights that, while correct, add no new information. These include paraphrasing the subtopic, contextual statements (\eg opening remarks, takeaway), as well as insights that are overly generic (and, thus, non-informative), or overly specific (\eg explanations, particular background information).\\
        \midrule 
        \sloppy\raggedright Instruction Inconsistency & Insights that are unrelated to the subtopic, redundant, or not shared, violating the conditions specified in the prompt: (1) focus on a subtopic, (2) create concise summaries, and, in the \subtopictrust setting, (3) focus on shared information.\\
        \midrule 
        \sloppy\raggedright Context Inconsistency & Insights that are subtly misrepresented, either by oversimplifying or exaggerating details. Examples include: \textit{overgeneralization}, where individual opinions are turned into broader claims (\eg individual's opinion reported in the predicted insight as a generic claim), or \textit{oversimplification}, where the model narrows details to limited contexts inappropriately (\eg when a document mentions both banks and government agencies, but the predicted insight only focuses on a specific bank).\\
        \midrule 
        \sloppy\raggedright Fabrication & Insights that contradict or are not supported by the information presented in the provided documents. In practice, fabrications are hard to identify and often manifest as slight changes in wording.\\
        \bottomrule
    \end{tabular}
\end{table*}
Understanding the nature and frequency of \llm hallucinations is crucial for evaluating the reliability of these models in practical applications. 
While prior work acknowledges failures in \llm outputs~\citep{summhay--laban-et-al-2024}, these failures are not well-studied in \mds.
To bridge this gap, we collect human annotations for 150+ \llm-generated summaries and proposed a taxonomy based on the recurring mistakes observed. 
Specifically, we manually inspect over 700 predicted insights across models, domains, and task focus. 
To assess whether predicted insights are \textit{faithful}, we also include the input documents in the annotation process.
However, as previously observed~\citep{chang2024booookscoresystematicexplorationbooklength}, analyzing long documents is quite challenging. 
Consequently, instead of analyzing multiple combination sizes, we limit our analysis to N=2, leaving the analysis of 2+ documents for future work. 
The main analysis is conducted by two authors of this paper (see Appendix \ref{app:sec:taxonomy} for further details on the annotation protocol). 
Finally, to ensure the robustness of the proposed error categories, we collect the category annotations of two additional authors across 50 examples spanning both domains for each category~\citep{chang2024booookscoresystematicexplorationbooklength}.
We discover that, on average, 80.71\% of the annotated insights are assigned the same label by 3 or more annotators.

We identify three common \llm errors, detailed in Table \ref{tab:hal-tax}: 
(1) focusing on high-level details (\textit{pedantic}), 
(2) failing to follow instructions (\textit{instruction inconsistency}), 
and (3) misrepresenting specific details (\textit{context inconsistency}).  
Occasionally, though less frequently, we also observe insights that deviate from the original input (\textit{fabrication}).
During the annotation process, we also encounter examples that fall into multiple categories, such as an overly generic insight (pedantic) related to a different subtopic (instruction inconsistency), suggesting the complexity associated with the errors observed in \mds settings.

Table \ref{tab:main:annotations:subtopic-and-trust:news} shows each error's frequency in the news domain. 
Overall, we find all \llms to be fairly faithful to input documents, as emphasized by the low rates of context inconsistency (9\%-37\%) and fabrication errors (0\%-9\%). 
The majority of the errors seems to be related to the generation of uninformative insights (pedantic) and unrelated subtopics (instruction inconsistency).
For instance, \shortgptfour and \shortqwen often include redundant and off-topic insights, while also adding coherence-enhancing insights such as takeaway statements. 
In \subtopictrust results, the rate of instruction inconsistency errors exceeds 70\% of all hallucinations. 
Of these, 80\% to 95.45\% (58.54\% to 67.24\% of all hallucinations) stem from insights not shared across documents, as required by the prompt.
\input{main/tables/hallucination-annotations-news-subtopic_and_trust}

\section{Mitigating Hallucination}
\label{sec:hallucination-mitigation}

With a clearer understanding of \llms hallucinations, we now explore whether we can mitigate them through simple post-processing heuristics. 
Focusing on \llms exhibiting higher rate of instruction inconsistent and pedantic errors, we investigate whether simple heuristics suffice to reduce incorrect insights (lower error rate) while preserving correct ones (maintaining recall).
In particular, we explore 4 mitigation methods that include output truncation (\texttt{top-k}), removal of redundant insights (\texttt{redundant}), subtopic paraphrases (\texttt{st-paraphrase}), and subtopic-unrelated (\texttt{st-unrelated}) (for technical details, see Appendix \ref{app:sec:mitigation:experiments}).
We apply each mitigation method to previously generated summaries in the 2-document \subtopic setting and report the absolute change in average F1-score.

Across mitigation methods, F1-score variation is minimal ($\pm$3\%), with a simple \textit{top-k} showing the most improvement (see Table \ref{tab:main:mitigation:gpt4-llama3-qwen}).
Additionally, \texttt{st-unrelated} and \texttt{redundant} have less impact than expected, despite targeting common errors in generated summaries. 
This may be due to poor performance of adopted \llm-based classifiers in the tested domains.
Even with a narrow focus (\subtopic, N=2), our findings highlight the complexity of reducing hallucinated errors in \llm-generated summaries within a \mds setting, calling for further research to address these challenges.
To minimize hallucination in \mds, future work could explore hierarchical approaches, where each document is summarized individually before combining the results into a final summary~\citep{chang2024booookscoresystematicexplorationbooklength,tang2024minicheckefficientfactcheckingllms}.
\begin{table}[tb]
    \small
    \centering
    \caption{\textbf{Absolute difference in average F1-score after applying four simple mitigation methods to summaries generated from two input documents (N=2)}. All methods show minimal impact on average F1-score ($\pm$3\%), with truncating summaries to the top 5 bullet-points being the most effective.}
    \label{tab:main:mitigation:gpt4-llama3-qwen}
    \begin{tabular}{ll ccc}
        \toprule
                                                        & \textbf{Strategy} & \textbf{\shortgptfour} & \textbf{\texttt{Llama 3.1}} & \textbf{\texttt{Qwen 2}} \\
        \midrule
        \multirow{4}{*}{\rotatebox{90}{news}}           & \texttt{top-5}     &  2.51\% &  1.69\% &  0.42\% \\ 
                                                        & \texttt{st-unrelated}  & -2.61\% & -1.49\% & -1.95\% \\ 
                                                        & \texttt{st-paraphrase}    & -1.19\% & -1.19\% & -0.86\% \\ 
                                                        & \texttt{redundant} & -0.49\% & -0.28\% & -0.98\%\\ 
        \midrule
        \multirow{4}{*}{\rotatebox{90}{conv}}           & \texttt{top-5}    &  2.28\% &  1.52\% &  1.52\% \\ 
                                                        & \texttt{st-unrelated} &  0.85\% &  0.85\% &  0.43\% \\  
                                                        & \texttt{st-paraphrase}    & -0.11\% & -0.46\% & -0.29\% \\ 
                                                        & \texttt{redundant} &  0.46\% & -0.08\% &  0.18\% \\ 
        \bottomrule
    \end{tabular}
\end{table}

%
%
\section{Related Work}
\label{sec:related_work}
\textbf{\emph{Multi-document summarization}} is a broad and versatile multi-document NLP task that consists of generating a summary from multiple source documents, including opinion/reviews \citep{angelidis-etal-2021-extractive-SPACE,iso-etal-2022-comparative}, scientific articles \citep{lu-etal-2020-multi-xscience,deyoung-etal-2021-ms,yang-etal-2023-oasum}, or news articles~\citep{fabbri-etal-2019-multi}.
Assessments of \llms capabilities can be carried using generic summarization benchmarks~(\citet{fabbri-etal-2019-multi,lu-etal-2020-multi-xscience,deyoung-etal-2021-ms,wolhandler-etal-2022-multi,huang-etal-2024-embrace}, \textit{inter alia}) or using focus-specific benchmarks, that resort to aspects~(\citet{angelidis-etal-2021-extractive-SPACE,hayashi-etal-2021-wikiasp,amar-etal-2023-openasp,yang-etal-2023-oasum}, \textit{inter alia}), queries~(\citet{kulkarni2020aquamuseautomaticallygeneratingdatasets,bolotova-baranova-etal-2023-wikihowqa,huang-etal-2024-embrace,chen-et-al-2024-MDCR,summhay--laban-et-al-2024}, \textit{inter alia}), or perspectives~\citep{naik-etal-2024-perspective}. 
However, these benchmarks mainly focus on overall performance metrics and rarely conduct in-depth error analysis, offering limited insight into \llms behavior and its relation to the input.

\noindent\textbf{\emph{Hallucination evaluation benchmarks}} have been proposed for single-document NLP tasks with the purpose of evaluating models' propensity to generate hallucinated content~\citep{huang2023surveyhallucinationlargelanguage}. 
These benchmarks are often designed to probe model errors, for instance, by testing knowledge of false beliefs and misconceptions~\citep{lin-etal-2022-truthfulqa,cheng2023evaluatinghallucinationschineselarge}, current events~\citep{vu-etal-2024-freshllms,kasai2024realtimeqawhatsanswer}, similar but incorrect statements~\citep{muhlgay-etal-2024-generating}, or even a models' ability to recover from nonsensical questions~\citep{pal-etal-2023-med}. 
While single-document hallucination benchmarks have been crucial in identifying and mitigating non-factual outputs in current \llms~\citep{touvron2023llama2openfoundation,yang2024qwen2technicalreport,openai2024gpt4,wei2022emergent}, 
hallucinatory behavior in multi-document tasks remains under-explored, especially how it varies with factors like document count or repeated or contradictory information.
%

\noindent\textbf{\emph{Automatic evaluation}}, while less reliable than human evaluation~\citep{kryscinski-etal-2019-neural,fabbri-et-al-2021-summeval}, is often sought for its scalability and efficiency in large-scale experiments, where human evaluation is impractical~\citep{krishna-etal-2023-longeval,chang2024booookscoresystematicexplorationbooklength,huang-etal-2024-embrace}.
To evaluate the grounding of LLM-generated text, specialized  entailment-based~\citep{laban-etal-2022-summac,tang2024minicheckefficientfactcheckingllms,goyal-durrett-2021-annotating} and question-answering-based~\citep{fabbri-etal-2022-qafacteval} metrics have been proposed. 
Recent work has found that, despite their higher cost, zero-shot prompting general-purpose LLMs performs as well as or better than specialized metrics when used for fact-checking while requiring no fine-tuning\citep{manakul-etal-2023-selfcheckgpt,tang2024minicheckefficientfactcheckingllms}.\footnote{
Similarly, prompting LLMs (with task specific prompts) has been shown to correlate better with human judgments than specialized models~\citep{liu-etal-2023-g,Farquhar2024}.}
Building on prior successes of LLM-as-a-judge approaches, we use an LLM-based metric with three demonstration examples~\citep{chang2024booookscoresystematicexplorationbooklength,summhay--laban-et-al-2024} to determine the information coverage between reference and predicted insights. 
While a variant of this metric has been rigorously validated~\citep{summhay--laban-et-al-2024}, we further confirm its reliability through human evaluations (see Sections \ref{sec:experiment_setup} and \ref{sec:error-taxonomy}), finding strong alignment with LLM judgments.

\section{Conclusion}
\label{sec:conclusion}
In this paper, we carry the first in-depth investigation of hallucinatory behavior of popular instruction-tuned \llms in a multi-document summarization task.
Through controlled experiments where we vary the number of documents and task focus, we find that up to 75\% of \llm-generated content is hallucinated.
More surprisingly, our results show that when prompting models to summarize information related to non-existing subtopics, models still generate plausible-sounding summaries in more than 20\% of the examples, with \shortchatgpt incorrectly generating summaries in 79.35\% of the samples. 
Subsequent manual analysis of 700+ insights reveals that, albeit \textit{faithful} to the input, a large fraction of insights is overly generic or inconsistent with the instructions.
To mitigate such errors, we experiment with simple \textit{post-processing} mitigation strategies but observe a trade-off with models' performances. 
Together, our results underscore the need for more effective approaches that systematically mitigate hallucinations in \mds. 

\section*{Limitations}
\label{sec:limitations}

While we design our evaluation protocol to ensure the reliability of our analysis, we acknowledge some limitations with our work.

\emph{Grounding evaluation in reference insights.}
To automate evaluation while keeping costs manageable, we compare LLM-predicted insights with the reference insights instead of using the documents. 
This assumes that the reference insights cover all relevant document content. This assumption is supported by the fact that SummHay documents were artificially generated from the reference insights and rigorously validated for completeness~\citep{summhay--laban-et-al-2024}.
However, this simplification could result in misclassifying a predicted insight as hallucinated, even if it is covered in the document.
Empirically, across 700+ insights, we find that most ``false hallucination'' results from models merging two high-level ideas, rather than omitting important information. 
Alternatively, future work could explore document-level~\citep{yin-etal-2021-docnli}, sentence-level~\citep{huang2023surveyhallucinationlargelanguage}, or fact-level~\citep{tang2024minicheckefficientfactcheckingllms} approaches to directly cross-check predicted insights with the documents themselves, rather than with reference insights.

\emph{Insights may not represent a single unit of information.} 
Underlying our evaluation protocol was the assumption that insights---both predicted and reference---represent a unique unit of information and ``are expected to mention a number, a date, or an entity'' and ``are independent of each other''~\citep{summhay--laban-et-al-2024}. 
However, we observe that the assumption does not always hold in practice (see examples in Appendix \ref{app:sec:dataset}), which has repercussions for the evaluation metric, which is produces a 1-to-1 coverage mapping between lists of insights. 
To address this, one solution could be to modify the evaluation prompt (listed in Figure \ref{fig:prompt:eval-prompt}) to output a list of ``coverage'' judgments (as opposed to a single JSON object). However doing so may introduce additional challenges for evaluation, such as anchoring effects for multi-attribute judgments~\citep{stureborg2024largelanguagemodelsinconsistent}. 
Alternatively, we could decompose the reference and predicted insights into smaller units of information~\citep{min-etal-2023-factscore,wanner-etal-2024-closer} and apply the metric on decomposed facts.

\emph{The use of \textit{information coverage} as a proxy for measuring similarity of meaning.} 
We acknowledge the pragmatic and semantic nuances involved in determining whether two insights have the same meaning. 
Linguistic phenomena, such as entailment, implicature, and presupposition~\citep{Dahlman_2021,jeretic-etal-2020-natural-IMPRESS,jiang-de-marneffe-2021-thinks,jiang-de-marneffe-2019-know} could prove useful in determining the semantic relationships between two sentences~\citep{goyal-durrett-2020-evaluating}.  
Alternatively, researchers have also explored other approaches~\citep{tang-etal-2023-understanding}, based on iterative question generation and response~\citep{wang-etal-2020-asking,scialom-etal-2021-questeval} or even based on prompting  \llm using a few demonstrations~\citep{chang2024booookscoresystematicexplorationbooklength,summhay--laban-et-al-2024}. 
In this work, similarly to previous work~\citep{huang-etal-2024-embrace,summhay--laban-et-al-2024}, we use the more abstract notion of \textit{coverage} (as opposed to other linguistic phenomena) due to its previous reports of high agreement with both experts and crowdsourcing annotators for abstractive summarization tasks. 

\emph{Hallucination Mitigation Trade-off.}
In this paper, we explore the efficacy of simple heuristics in the mitigation of hallucinations across \llms, which were designed to tackle specific issues with the \llm-generated summaries, including redundancy~\citep{xiao-carenini-2020-systematically}.  
However, we observe limited effect in \llms' hallucination rate in practice, raising important questions about the efficacy of these methods.
One potential explanation is that the \llm-based classifiers, used to detect these errors, perform poorly in the specific domains tested in this work. 
Moreover, we observe that most insights (both reference and predicted) are rarely atomic, containing a mixture of redundant and non-redundant pieces of information, making it difficult to detect using binary judgments. Future work could explore the use of more fine-grained atomic evaluation methodologies~\citep{min-etal-2023-factscore,Chen-2023-FELM-NeurIPS,tang2024minicheckefficientfactcheckingllms}. 

\section*{Acknowledgments}
We thank the members of Megagon Labs and UCI NLP for their valuable feedback and fruitful discussions.

\bibliography{main}

\appendix

\begin{figure*}[tb]
    \centering
     \includegraphics[width=\columnwidth]{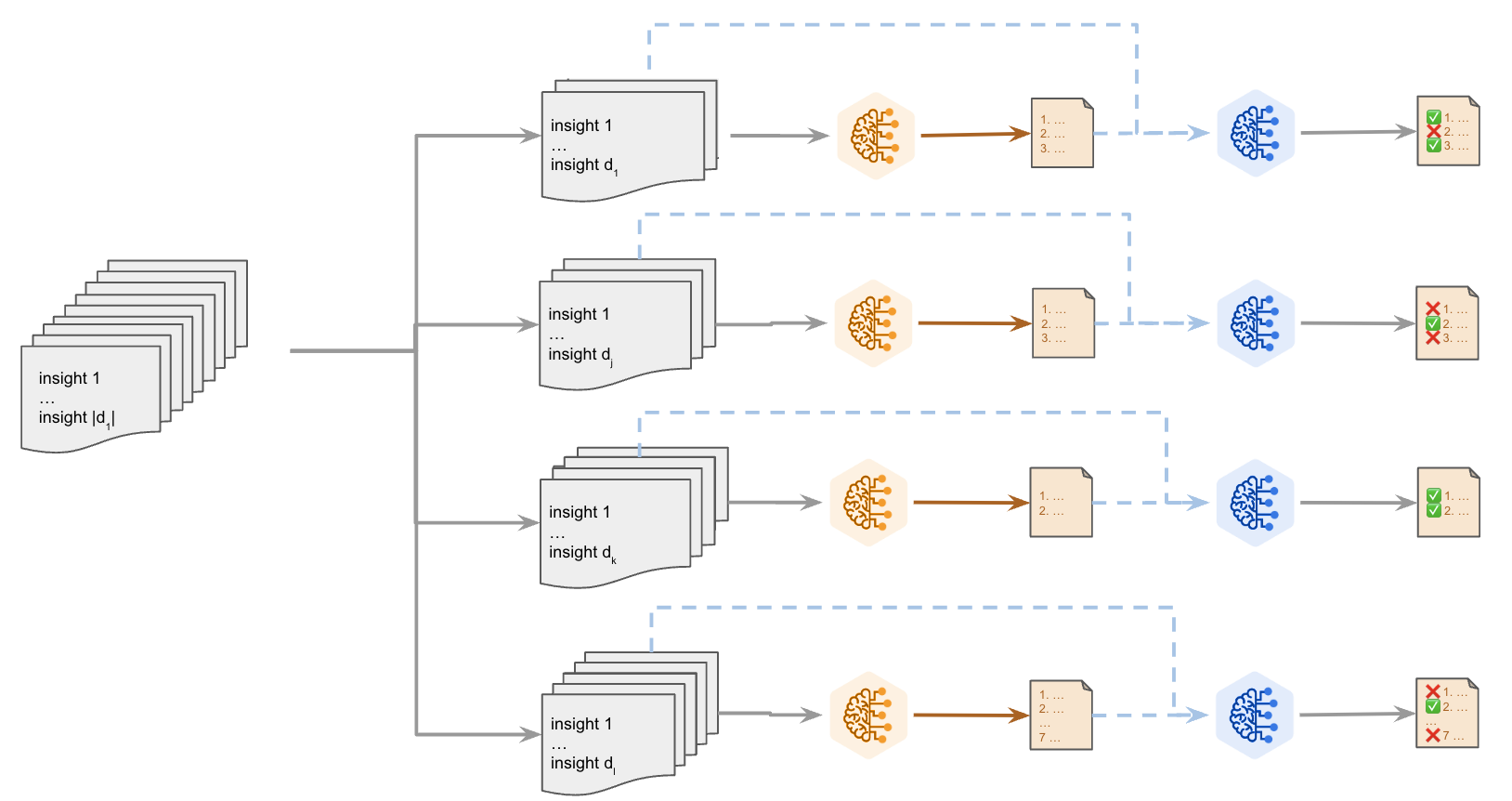}
    \caption{\textbf{Proposed evaluation protocol to investigate hallucinations in \llms as a function of number of documents and task focus}. Given a corpus of documents with insight-level annotations, we craft combinations of $N$ documents such that part of the information about the queried subtopic is present in 2+ documents.}
    \label{fig:evaluation-methodology}
    \vspace{-1em}
\end{figure*}

\section{Dataset Details}
\label{app:sec:dataset}

This section describes the dataset card and the access to the the datasets used. We also report the statistics of the benchmarks we created and that we base our analysis on.

\subsection{Original Datasets} 
\label{app:ssec:original-datasets}

To the best of our knowledge, there is still no \mds benchmark focused on the investigation of hallucinations. 
To this end, we leverage an existing dataset---SummHay\citep{summhay--laban-et-al-2024}---that provides fine-grained level annotations about the relevant information presented in each insight.
Originally proposed to evaluated \llms' performance in long-context summarization, SummHay puts forward two synthetic datasets spanning different domains: news and conversation, which we refer to as \newsdataset and \convdataset, respectively. 
Each dataset comprises 500 documents, uniformly distributed across 5 different topics (see list of topics in Table \ref{tab:app:topics-summhay}). 
Each document is carefully crafted using \shortgptfour to ensure that facts (or \textit{insights}) are repeated across 6+ documents within the same topic. 
To this end, the authors begin by prompting \shortgptfour to generate a list of candidate subtopics that are unique and expandable into more than 3 distinct insights.
Each subtopic is then automatically validated to ensure that no subtopics overlap thematically and to verify that there are at least 3 insights that are specific to that subtopic and not other.
For additional details on the dataset creation, refer to the original paper.

\input{appendix/tables/topics_summhay}
\input{appendix/examples/ref-insights}

\subsection{Evaluated Datasets.}
\label{app:ssec:evaluated-datasets}

One of our driving research questions was to understand how models' predisposition to hallucinate changed with increasing number of documents. 
To this end, we repurpose the SummHay dataset by manipulating the insight-level annotations associated with each document. 
In particular, for every topic (each containing 100 documents), we create combinations of $N$ documents s.t. when paired with a subtopic $q$, the resulting combination is guaranteed to have at least 2 subtopic-related insights that are shared across 2 or more documents. 
We perform this for various combination sizes $N=\{2,3,4,5,10\}$ for all 10 topics, resulting in two benchmarks that we use in our evaluations. 
The statistics of the benchmarks are summarized in Tables \ref{tab:dataset-statistics-news} and \ref{tab:dataset-statistics-conversation}, including the input length\footnote{The length of the combinations is estimated using \texttt{tiktoken}'s encoder \texttt{cl100k\_base} to encode all documents in a combination.} (\texttt{total length}), number of shared subtopics (\texttt{\# shared subtopics}), and shared insights (\texttt{\# shared insights}) 
From these benchmarks, we sample 500 combinations for each $N$, totalling 2.5k examples per domain. 
\input{appendix/tables/dataset_statistics/news-domain}
\input{appendix/tables/dataset_statistics/conversation-domain}

\section{Prompt Selection}
\label{app:sec:prompt-selection}

This paper outlines the reasoning and proposed modifications to the prompts used in the SummHay paper. 
The prompts used to report the results in the main paper are: 

\begin{itemize}
\item Figure \ref{fig:prompt:summary-news:subtopic}, used in the news, \subtopic setting;
\item Figure \ref{fig:prompt:summary-news:subtopic+trustworthy}, used in the news, \subtopictrust setting;
\item Figure \ref{fig:prompt:summary-conv:subtopic}, used in the conv, \subtopic setting;
\item Figure \ref{fig:prompt:summary-conv:subtopic+trustworthy}, used in the conv, \subtopictrust setting;
\item Figure \ref{fig:prompt:eval-prompt}, used to parameterize the \llm-evaluator and measure information coverage.
\end{itemize}

\begin{figure*}[tb]
    \centering
    \includegraphics[width=\textwidth]{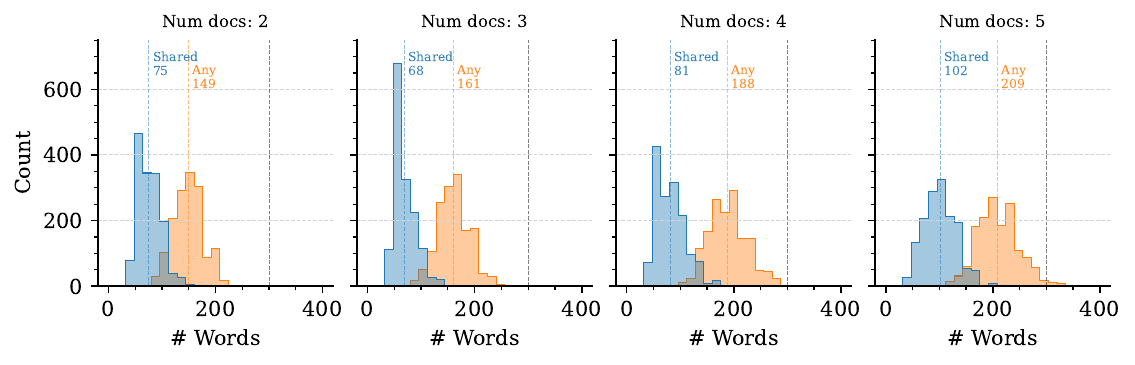}
    \caption{\textbf{Distribution of the number of words of the reference summaries for the proposed evaluation benchmark in the news domain}. The length of reference summaries is computed by concatenating ground truth reference insights for the different examples in the evaluation benchmarks and counting the number of words separated by whitespaces. \texttt{Any}, and \texttt{Shared} refer to the \subtopic and \subtopictrust settings, respectively. We observe that, across the different document combinations, the number of words (including punctuation) is fewer than the limit of 300 words specified in the prompt proposed by \citet{summhay--laban-et-al-2024}. Thus suggesting that 300 words is more than enough to faithfully summarize information within the provided documents.}
    \label{fig:histplot:ground-truth-num-words}
\end{figure*}

Early on in our experiments, we explored a modified version of the news summarization prompt proposed in the SummHay paper~\citep{summhay--laban-et-al-2024}. 
Since our goal was to investigate hallucinations in \mds (and not in model's capability to cite different sources), we remove the instructions related to the citation task, therefore avoiding potential anchoring effects that could be associated with performing multiple tasks in a single shot~\citep{stureborg2024largelanguagemodelsinconsistent}.
Additionally, because we are interested in analyzing model behavior in scenarios where the number of relevant insights in the input documents is unknown, we do not instruct the model on the ideal summary length. 
Instead, in our earlier experiments, to avoid biasing evaluation based on the length of the summaries~\citep{liu-etal-2023-revisiting,guo-vosoughi-2023-length}, we explore instructing the model to restrict the number of words used in the summaries. 
To determine this number, we analyse the ground truth summaries (concatenation of reference insights for each dataset) and find 300 words to be a reasonable upper bound over the reference insights (see Figure \ref{fig:histplot:ground-truth-num-words}), therefore allowing for some semantic and lexical variations in \llm summaries. 

\begin{figure}[tb]
    \centering
    \includegraphics[width=\linewidth]{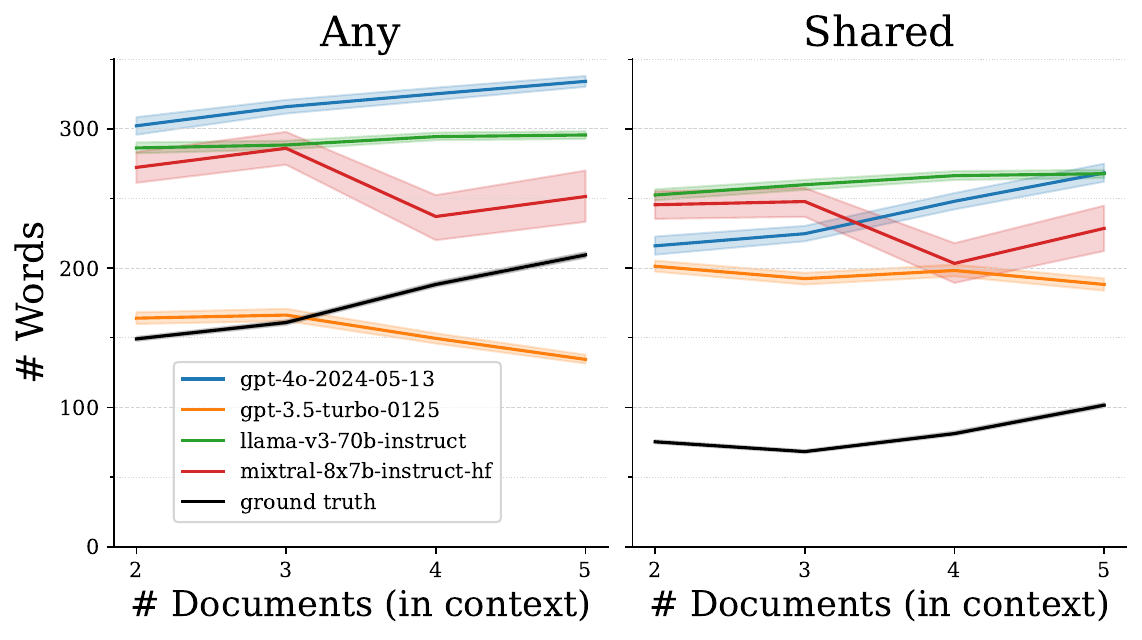}
    \caption{\textbf{Number of words in summaries as a function of the number of input documents (news domain)}. On the left, we observe the length of LLM-generated summaries when instructed to summarize insights in the \subtopic setting. On the right, we observe the length of LLM-generated summaries when instructed to summarize shared subtopic-related insights (\subtopictrust setting). In general, we observe a large gap between the total length of ground truth summaries (black solid line) and the LLM-generated ones, suggesting that models attempt to generate 300 words-long summaries even when the relevant information can be condensed in using fewer words.}
    \label{fig:lineplot:length-models-conditioned300}
    \vspace{-1em}
\end{figure}

\paragraph{Ablation: Instructing \llms to adhere to a fixed number of words. }
Using the modified prompt, we instruct 4 different \llms to generate 250 summaries for different number of document combinations and investigate the properties of the generated summaries, including their sentence length. 
To this end, we decompose the summaries into their words and punctuation using the \texttt{word\_tokenize} from the \texttt{nltk} Python library\footnote{\url{https://www.nltk.org/api/nltk.tokenize.html\#nltk-tokenize-package}} and report the number of generated words. 
Lengthwise, with the exception of \texttt{gpt-3.5-turbo-0125 (Any)}, we observe a large discrepancy (up to 200 words difference) between \llm-generated summaries and ground truth summaries, especially in cases involving fewer input documents (see Figure \ref{fig:lineplot:length-models-conditioned300}). 
One potential reason for the observed discrepancy is that models attempt to satisfy the length condition (of 300 words) and, in doing so, generate overly verbose summaries. 
Such verbosity may negatively impact models' ability to digest information and increase the models' hallucination rate. 

\begin{figure}[tb]
    \centering
    \includegraphics[width=\linewidth]{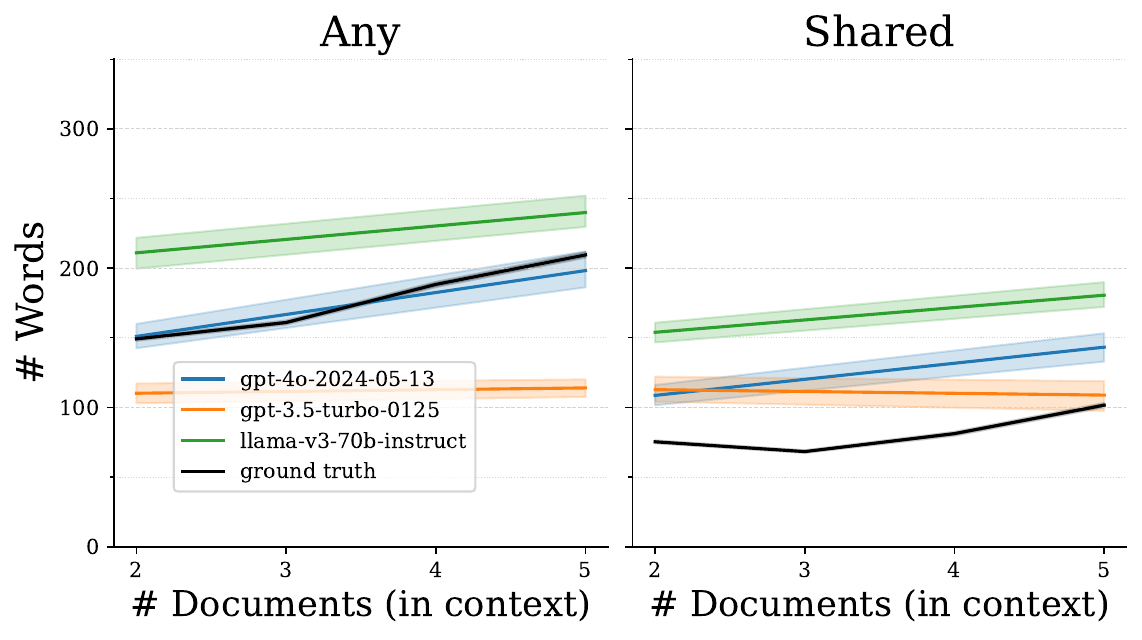}
    \caption{\textbf{Number of words in summaries as a function of the number of input documents (news domain)}. The summaries are generated by instructing the models on the generation of \textit{clear and concise} summaries instead of \textit{no longer than 300 words in total}. We observe smaller discrepancies between ground truth summaries and LLM-generated summaries. Compared to Figure \ref{fig:lineplot:length-models-conditioned300}, LLM-generated summaries are shorter by up to 150 words, especially for fewer documents. These results suggest that specifying the length of the summaries may bias the model to produce unnecessarily verbose summaries.}
    \label{fig:lineplot:length-models-noconditioned300}
    \vspace{-1em}
\end{figure}

\paragraph{Ablation: Instructing \llms to be succinct. }
In many practical applications, it is difficult to exactly antecipate how many words (or insights) would be enough to summarize the relevant information, since it may depend on various factors, including the amount of information in the input (\eg number, length, and diversity of documents), as well as the degree of specificity of the summary (\eg depth vs more general insights). 
Consequently, we opt for instructing models with a more nuanced length restriction that emphasizes the clarity and conciseness of the generated summary: \textit{Your summary should be concise and clear.}. 
Empirically, we found that replacing the fixed-length instruction led to significant reductions (by 50 to 150 words) in the length of \llm-generated summaries, leading to summaries that length-wise are more aligned with the reference summaries (see Figure \ref{fig:lineplot:length-models-noconditioned300}). 
Based on these findings, we replace the fixed-length condition in the prompt with this more nuanced instruction. 

\paragraph{Addressing mispecifications in the input. }
Finally, to provide some format to the summaries and limit potential confounding aspects related to the bullet-point or the inability to identify subtopic-related insights in the prompt, we include two additional instructions in the prompt: \textit{Represent each bullet-point using "-".} and \textit{If you do not find related insights, write "No insights found" and nothing else.}. 
\input{appendix/prompts/summary_prompt}

\input{appendix/prompts/evaluation_prompt}

\section{Model Access Details}
\label{app:sec:model-access-details}

This section describes the model card, access endpoints, and hyperparameter configurations used for each model.

\paragraph{Model Access. } 
We use the official OpenAI API\footnote{\url{https://platform.openai.com/docs/quickstart}} to access three \llms, namely \fullchatgpt, \fullgptfour, and \fullgptmini. 
To access \fullgemini we use Google's Generative AI API\footnote{\url{https://ai.google.dev/gemini-api/docs/quickstart?authuser=2&lang=python}}.
The remaining models, which include \fullllama and \fullqwen, are accessed through the Fireworks API\footnote{\url{https://fireworks.ai/models}}. 
Early on we also used the Fireworks API to evaluate \llms, including Llama-3 (70B) (\texttt{llama-v3-70b-instruct}), and Mixtral (\texttt{mixtral-8x7b-instruct-hf}) but found numerous artifacts in their generations and, for that reason, excluded them from our analysis.
For instance, \texttt{Llama-3 (70B)} generates preambles before generating list items (\eg \textit{Here is a summary ...}, \textit{Here are the insights regarding ...}) but also postambles in 43\% of its generations. In the postambles \texttt{Llama-3 (70B)} either re-iterate the instructions (\eg \textit{Note: These insights are summarized from both Article 1 and Article 2}, \textit{These insights ... and are summarized in a concise and clear manner.}) or summarizes the listed bullet-points (\eg \textit{These insights highlight the issues with Twitter's complex tweet recommendation algorithm ..., which is a critical step in improving the platform's user experience and rebuilding trust with its users.}). Conversely, \texttt{mixtral-8x7b-instruct-hf}'s generations either end abruptly, leading to incomplete summaries (\eg \textit{Foot Locker aims to ensure that approximately 40\% of its revenue}, \textit{Twitter, under Elon Musk's leadership, has announced plans to}), or conclude with the total word count (\eg \textit{(Word count: 241)}, \textit{(Note: The word count is exactly 200 words without the headings.)}). 
Throughout our experiments, we explicitly focus on the evaluation of larger models due to their superior instruction following capability. 
All experiments were carried in 2024, with the main experiments (in Section \ref{sec:experimental-results:investigation-results}) being conducted between July 1st to September 5th, and the mitigation experiments (described in Section \ref{sec:hallucination-mitigation}) being conducted between September 10th and October 12th.

\paragraph{Summary generation. }
The summaries generated in this paper were all generated using the following configurations: \texttt{temperature=1}, \texttt{top\_p=0.9}, and \texttt{max\_tokens=800}.

\paragraph{LLM-based evaluation. } 
Automatic evaluation is conducted using \texttt{gpt-4o-mini-2024-07-18} with parameters \texttt{n=1}, \texttt{temperature=0}, \texttt{top\_p=1}, and \texttt{response\_format: \{ "type": "json\_object"\}}. 
In Appendix \ref{app:sec:automatic-metric-validation}, we validate that the metric correlates strongly with human judgements of correctness.

\section{Automatic Metric Validation}
\label{app:sec:automatic-metric-validation}

In the main paper, we adopt an LLM-as-a-judge approach~\citep{zheng2023judgingllmasajudgemtbenchchatbot} to determine whether the predicted insights \textit{faithfully} cover the information in the input. 
Our evaluation approach is based on the metric proposed in the SummHay's paper~\citep{summhay--laban-et-al-2024}, which comprehensively compared the accuracy and human-alignment of different \llms as evaluators, including \texttt{gpt-4o}~\citep{openai-gpt-4o-blogpost}, \texttt{Gemini-1.5-pro}~\citep{geminiteam2024gemini}, and \texttt{Claude}~\citep{claude2024-blogplost}. They find that among models they evaluate, \texttt{Gemini-1.5-pro} and \texttt{gpt-4o} were found to be positively correlated with human-level annotations. 
In this paper, however, we adopt \texttt{gpt-4o-mini-2024-07-18} as a more cost-efficient but still competitive evaluator.\footnote{\texttt{gpt-4o-mini-2024-07-18} has been shown to be competitive with \texttt{gpt-4o} across various math, coding, and instruction-following benchmarks alike~\citep{openai-gpt-4o-mini-blogpost,chiang2024chatbotarenaopenplatform}.} 
To ensure that replacing the evaluator model does not negatively impact evaluation quality \citep{summhay--laban-et-al-2024}, we investigate the agreement between the two \llms across 45.3k reference insights, spanning 4 summarization models and 2 distinct prompts.
Given the complex nature of the evaluation process, our analysis is decomposed in terms of \textit{linking agreement}---focused on whether models agree on which predicted insight (if any) covers the information of the reference insight---and \textit{coverage agreement}---measuring to which extent the models agree on the coverage label (\nocov, \partcov, \fullcov).

\paragraph{Model Coverage Agreement. }
Among the matching examples, \texttt{gpt-4o} and \texttt{gpt-4o-mini} exhibit strong inter-annotator agreement, achieving a Cohen's Kappa of 0.84 and a Kendall Tau correlation coefficient of 0.94 \citep{McHugh2012-interannotator-reliability,kendall-tau}. 
In practice, the two models agree on the coverage labels for about 91\% of these examples (see Table \ref{tab:automatic-eval-disagreement-label-coverage-breakdown}).  
\begin{table}[tb]
    \centering
    \caption{\textbf{Breakdown of information coverage labels for the 43.2k examples where \texttt{gpt-4o} and \texttt{gpt-4o-mini} agree on the candidate insight.} The two models output the same coverage label in 91\% of the examples, with \texttt{gpt-4o-mini} being more optimistic than \texttt{gpt-4o} in the cases where they disagree.}
    \label{tab:automatic-eval-disagreement-label-coverage-breakdown}
    \begin{tabular}{llrc}
        \toprule
        \texttt{gpt-4o} & \texttt{gpt-4o-mini} & Counts & Frequency \\
        \midrule
        full & full & 13084 & 30.24\\
        full & partial & 18 & 0.04 \\
        partial & full & 3877 & 8.96\\ 
        partial & partial & 2255 & 5.21 \\
        no & no & 24029  & 55.55\\
        \bottomrule
    \end{tabular}
    \vspace{-1em}
\end{table}

\paragraph{Model Linking Agreement. } 
Our results show that the models' linking predictions match in about 95.5\% of the examples, suggesting strong linking agreement between the two models.
Out of the 2049 mismatching examples, we find that \texttt{gpt-4o-2024-05-13} and \texttt{gpt-4o-mini-2024-07-18} assign the same coverage label to 24.35\% of them (see Table \ref{tab:automatic-eval-linking-disagreement-label-coverage-breakdown}). 
Through manual inspection of 50 such examples, we find that the candidate insights exhibit high semantic overlap or complementarity, which is not well captured by the current evaluation process---only 1 candidate insight can be attributed to the reference insight. 
When comparing the win rate of the two models, we find that humans agree with both models 55\% times and disagree with both models about 10\% of the times. 
We also note that while \texttt{gpt-4o-2024-05-13} favors predicted insights with higher lexical overlap with the reference insight, humans prefer the answers from \texttt{gpt-4o-mini-2024-07-18} (over \texttt{gpt-4o-2024-05-13}) 30\% of the times. %

Next, we consider the cases where \texttt{gpt-4o-2024-05-13} suggests a greater information coverage of the reference insight than \texttt{gpt-4o-mini-2024-07-18}---representing 39.09\% of the examples with linking disagreement.
Again, we manually inspect 50 such examples and find that humans prefer \texttt{gpt-4o-mini-2024-07-18} over \texttt{gpt-4o-2024-05-13} in 60\% examples. 
One potential reason for the high disagreement is due to the fact that the predicted insights only match the information content of reference insights at a high-level, missing important details or adding irrelevant information, therefore compromising the usefulness of predicted insights in practice.
As a result, humans may prioritize insights that cover part of the details within the reference insight (as opposed to insights focusing on the higher-level information).
Similar human trends are observed when analyzing cases for which \texttt{gpt-4o-mini-2024-07-18} attributes higher information coverage than \texttt{gpt-4o-2024-05-13}. 
Across 50 manually annotated examples, humans prefer \texttt{gpt-4o-mini}'s outputs about 46\% of the times \textit{versus} 28\% of examples where they prefer \texttt{gpt-4o} outputs, having 18\% of examples where both are equally preferred.
\begin{table}[tb]
    \centering
    \caption{\textbf{Breakdown of information coverage labels for the 2049 non-matching examples}. In 24.35\% of the mismatches, the models predict the same coverage label (\textit{full}, \textit{partial}, \textit{no}) for different insights. \texttt{gpt-4o-mini} is more pessimistic than \texttt{gpt-4o} in 39.09\% of the examples and more optimistic in the remaining 36.56\%.}
    \label{tab:automatic-eval-linking-disagreement-label-coverage-breakdown}
    \begin{tabular}{llrr}
        \toprule
        \texttt{gpt-4o} & \texttt{gpt-4o-mini} & Counts  & Relative Freq \\
        \midrule
        full & full & 344 & 16.79 \\
        full & partial & 1 & 0.05\\
        partial & full & 336 &  16.40 \\
        partial & partial & 155 & 7.56 \\
        partial & no & 800 & 39.04 \\
        no & full & 54 & 2.64 \\
        no & partial & 359 & 17.52 \\
        \bottomrule
    \end{tabular}
    \vspace{-1em}
\end{table}

\section{Additional Results}
\label{app:sec:additional-results}

This section constitutes additional results that are complementary to the main paper.
\begin{itemize}
    \item Appendix \ref{app:ssec:impact-cov-labels} discusses the impact of coverage labels in the reported metrics;
    \item Appendix \ref{app:ssec:number-insights} provides insights about the length of the generated summaries;
\end{itemize}

\subsection{Impact of Coverage Labels}
\label{app:ssec:impact-cov-labels}

In Section \ref{ssec:methodology:evaluation}, we described the \llm-based metric that we use to automatically determine the \textit{faithfulness} of the generated summary. 
By design, our metric outputs three coverage labels (\nocov, \partcov, \fullcov). 
But to be able to compute the recall and hallucination rates, we must covert these into a single correctness label (\textit{correct/incorrect}). 
In the original paper~\citep{summhay--laban-et-al-2024}, the authors sidestep this conversion by computing a \textit{coverage score} (that is similar to recall) but is computed as follows: 
\begin{quote}
\textbf{Coverage Score}: For each insight, the summary receives a score of 100 for full coverage, 50 for partial coverage, and 0 otherwise. The final coverage score of a summary is the average coverage on all the insights of the subtopic, such that it ranges from 0 to 100.
\end{quote}

\begin{figure*}[tb]
    \centering
    \vspace{-1em}
    \begin{subfigure}[b]{\columnwidth}
        \centering
        \includegraphics[width=\columnwidth]{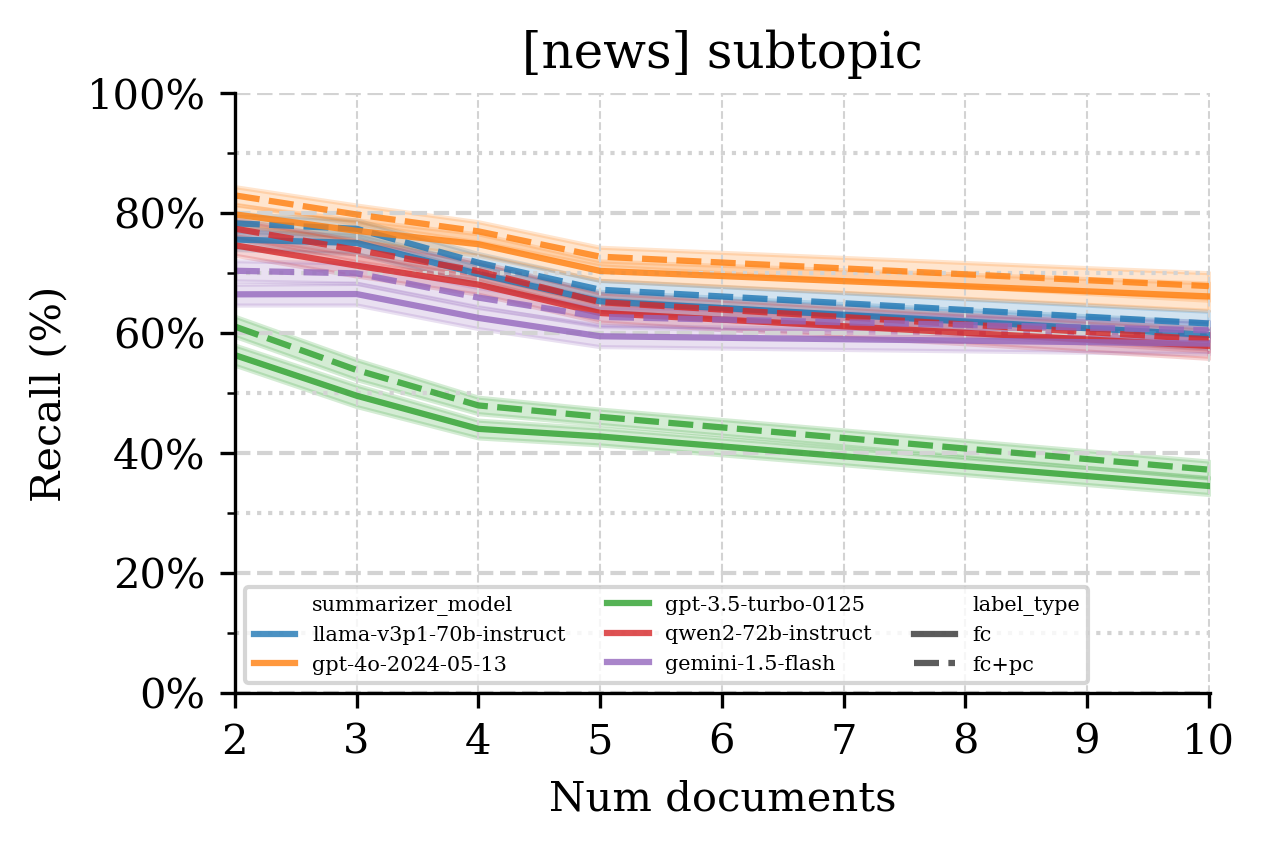}
        \label{fig:app:impact-cov-label:recall:subtopic:news}
    \end{subfigure}
    \begin{subfigure}[b]{\columnwidth}
        \centering
         \includegraphics[width=\columnwidth]{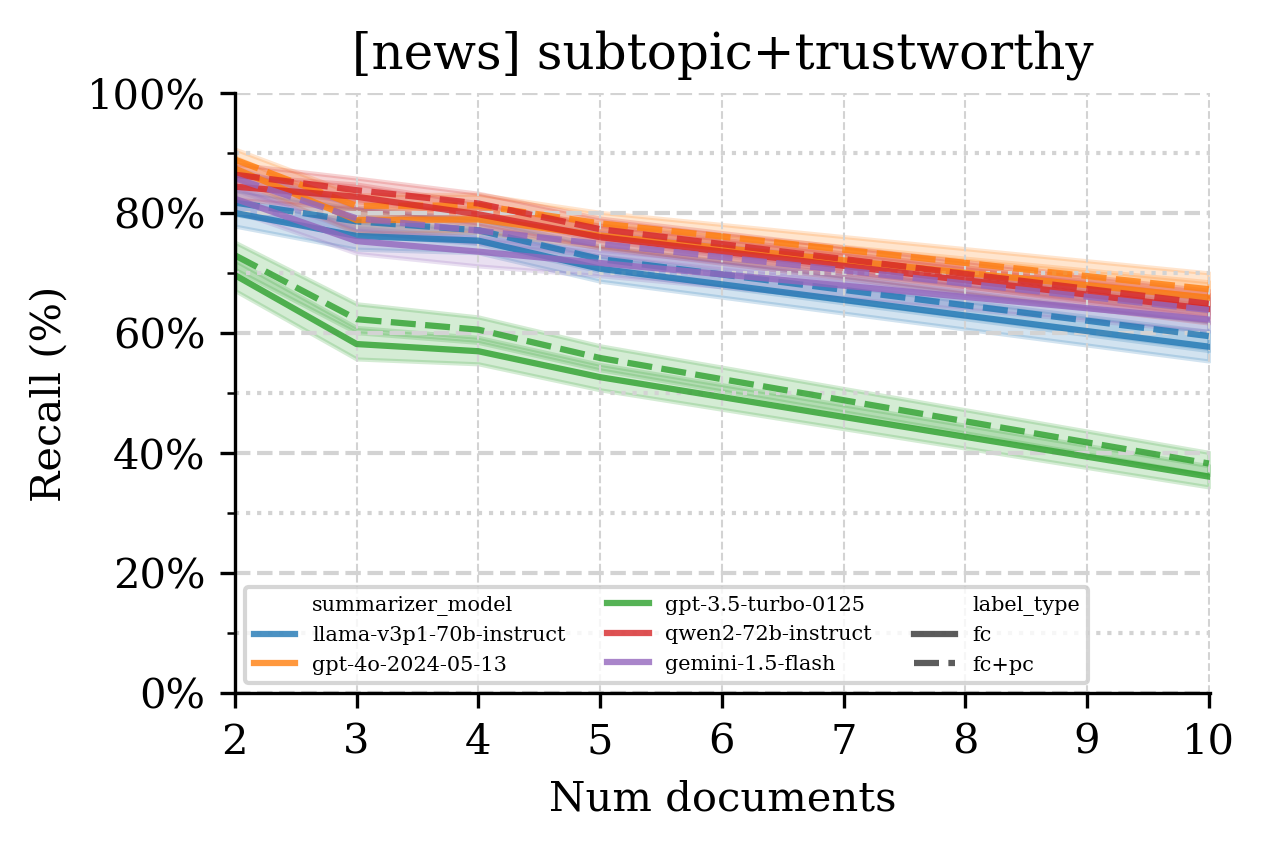}
        \label{fig:app:impact-cov-label:recall:subtopic+shared:news}
    \end{subfigure}
    \caption{\textbf{Impact of coverage labels in the macro-recall metric for the \newsdataset across both prompt settings}. Overall, we observe minimal differences ($<5\%$) across evaluated \llms, number of documents, and prompt settings.}
    \label{fig:app:impact-cov-label:recall:news}
    \vspace{-1em}
\end{figure*}

\begin{figure*}[tb]
    \vspace{-1em}
    \centering
    \begin{subfigure}[b]{\columnwidth}
        \centering
        \includegraphics[width=\columnwidth]{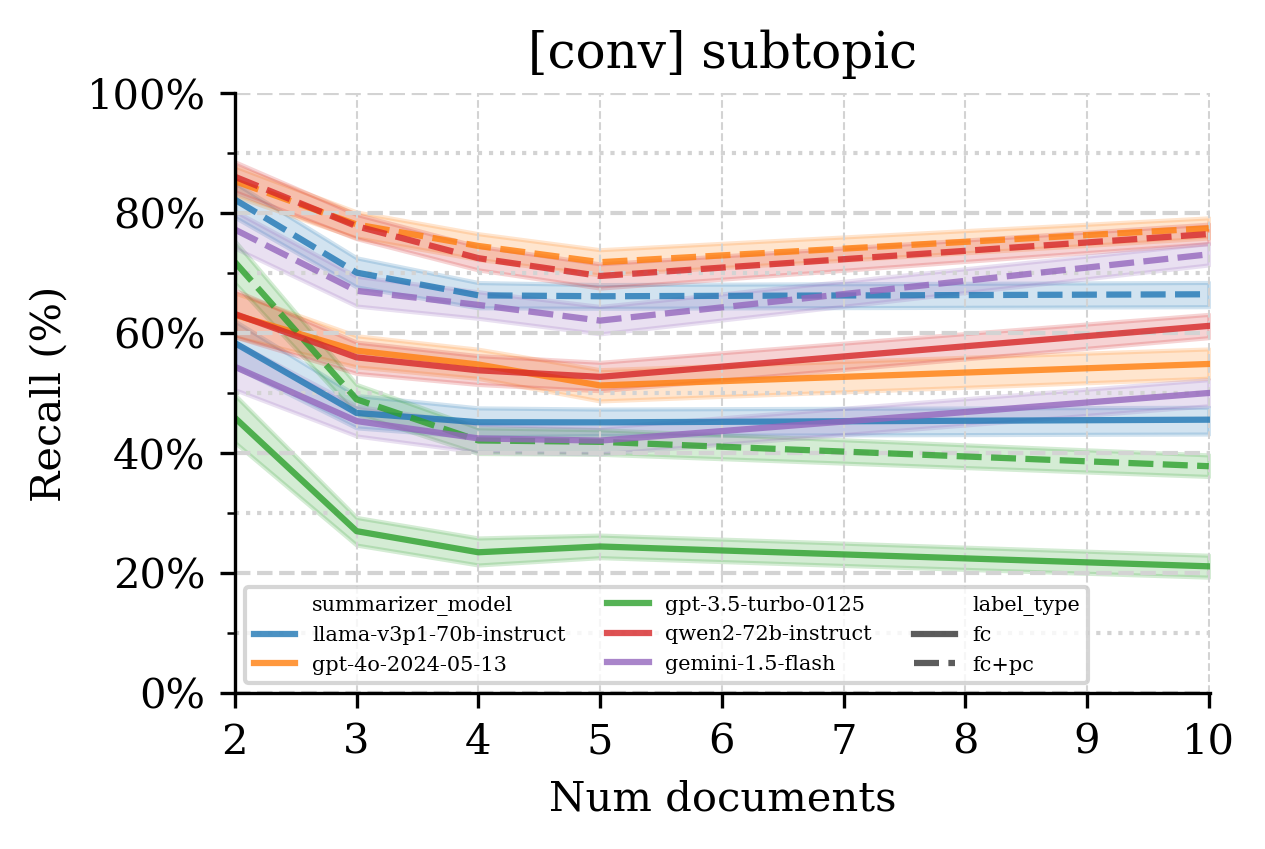}
        \label{fig:app:impact-cov-label:recall:subtopic:conv}
    \end{subfigure}
    \begin{subfigure}[b]{\columnwidth}
        \centering
         \includegraphics[width=\columnwidth]{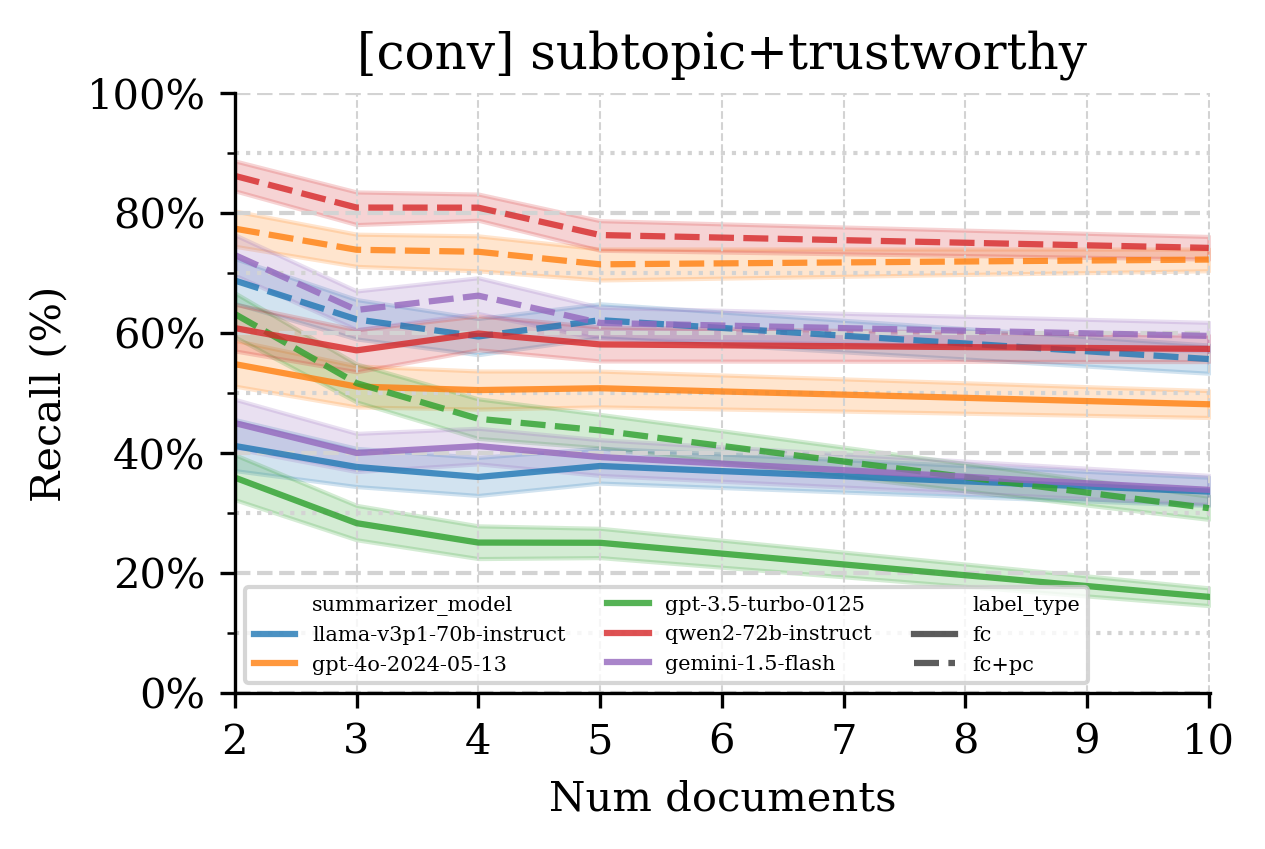}
        \label{fig:app:impact-cov-label:recall:subtopic+shared:conv}
    \end{subfigure}
    \caption{\textbf{Impact of coverage labels on recall measures in the \convdataset}. Unlike in the news domain, we observe large disparities of up to 30\% percentage points depending on whether we consider partial coverage (fc+pc) or not (fc). We hypothesize that the higher fraction of \partcov labels is associated with the fact that insights in the conversation domain are more contextual and are less entity-centric than the news domain and, as a consequence, make it more difficult to assess but also to summarize.}
    \label{fig:app:impact-cov-label:recall:conv}
    \vspace{-1em}
\end{figure*}

Instead, we adopt an optimistic approach and report the best-case scenario. In other words, by assuming that both \partcov and \fullcov insights are correct, we get a ceiling on the model's performance in the best case scenario, \eg  if a model achieves 20\% recall or 60\% error rate, we immediately know that in a more conservative scenario, models will exhibit lower performance. 
Note, however, that in practice this difference between \partcov and \fullcov is only meaningful if a \partcov represent a large fraction of the coverage labels assigned by our metric.
Figure \ref{fig:app:impact-cov-label:recall:news} shows that indeed the gap between the two is minimal for the news domain. 
Conversely, Figure \ref{fig:app:impact-cov-label:recall:conv} illustrates an up to 30\% points difference when considering \partcov correct (fc+pc) or incorrect (fc).

\subsection{Analysis of the Number of Insights}
\label{app:ssec:number-insights}

\begin{figure*}[tb]
    \centering
    \begin{subfigure}[b]{\columnwidth}
        \centering
        \includegraphics[width=\linewidth]{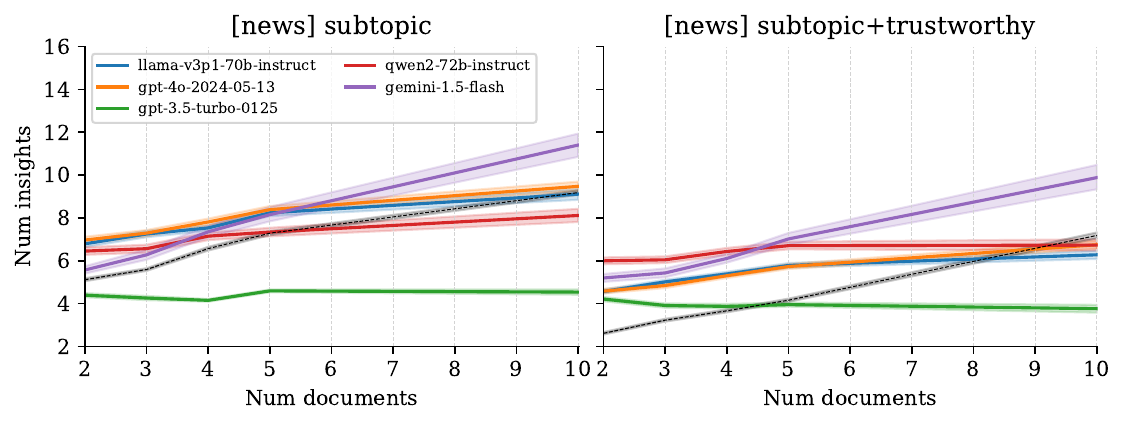}
        \caption{news}
        \label{fig:app:num-insights:news}
    \end{subfigure}
    \begin{subfigure}[b]{\columnwidth}
        \centering
        \includegraphics[width=\linewidth]{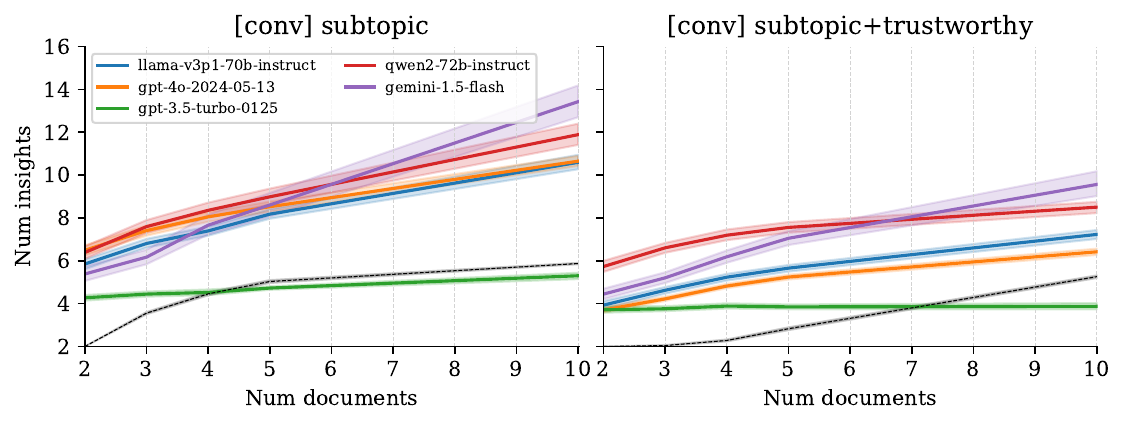}
        \caption{conversation}
        \label{fig:app:num-insights:conv}
    \end{subfigure}
    \caption{\textbf{Number of predicted insights reported for the the proposed evaluation benchmarks}. Each line represents the average number of insights, with shaded region representing the 95\% confidence interval. Despite being less perceptible in the news domain, we observe systematic differences between the two prompt settings, suggesting that models exhibit different behavior when instructed to focus on the shared insights.}
    \label{fig:app:num-insights}
    \vspace{-1em}
\end{figure*}

As previously mentioned (in Appendix \ref{app:sec:prompt-selection}), we do not to condition models to generate summaries with a fixed number of insights. 
As a consequence, and because of the different training dynamics and fine-tuning procedures associated with different \llms, we expect models to generate summaries with different lengths.
Figure \ref{fig:app:num-insights} shows the average number of insights for the evaluated benchmarks. We observe that, with the exception of \fullchatgpt, models tend to generate longer summaries.

\begin{figure*}[tb]
    \centering
    \begin{subfigure}[b]{\columnwidth}
        \centering
        \includegraphics[width=\linewidth]{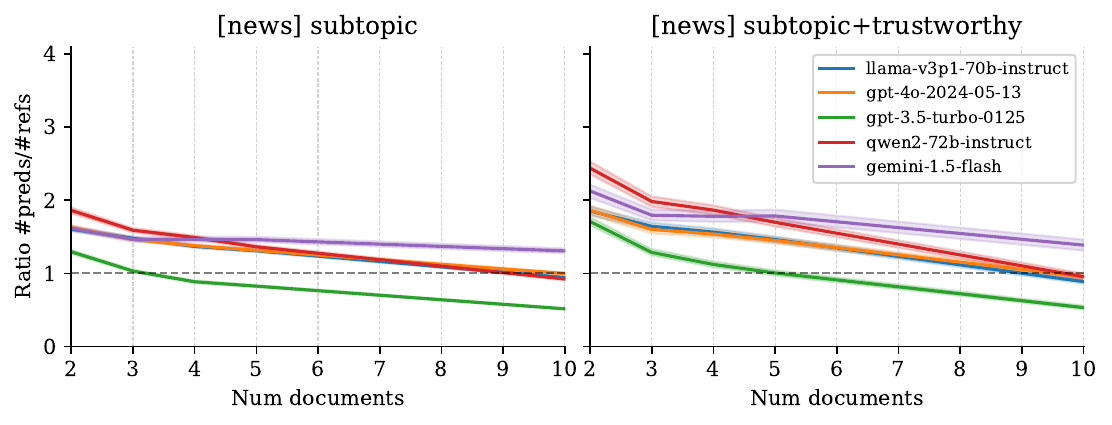}
        \caption{news}
        \label{fig:app:ratio-insights:news}
    \end{subfigure}
    \begin{subfigure}[b]{\columnwidth}
        \centering
        \includegraphics[width=\linewidth]{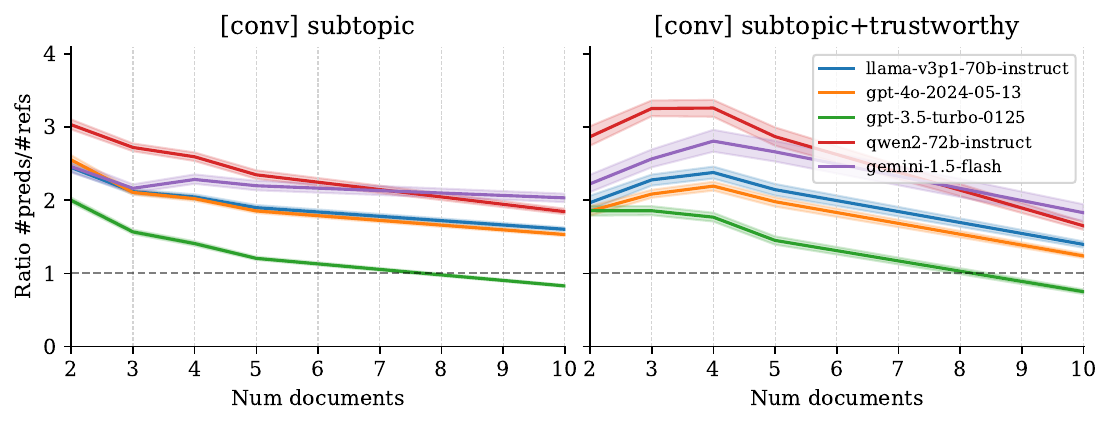}
        \caption{conv}
        \label{fig:app:ratio-insights:conv}
    \end{subfigure}
    \caption{\textbf{Ratio of predicted insights to reference insights reported for two datasets}. Each line represents the average number of insights, with shaded region representing the 95\% confidence interval. A ratio greater than 1 indicates the model predicted more insights than those referenced and, thus, indicates larger propensity to make mistakes (or possibly more redundant information). A ratio less than 1 signifies fewer predicted insights, indicating the model fails to identify some of the relevant information in the input documents.}
    \label{fig:app:ratio-insights}
    \vspace{-1em}
\end{figure*}

Figure \ref{fig:app:ratio-insights} shows the ratio of predicted-to-reference insights for the summaries generated for the two evaluated domains---news and conversation---as the number of input documents increases. By observing this ratio we can examine each models' propensity to over- or under-generate insights.
The first observation is that model behavior varies considerably depending on the prompt setting (\subtopic or \subtopictrust) and the domain.
Secondly, we observe that all models tend to over-estimate the number of relevant insights in the input documents when processing lower number of documents (N<3).
In general, we find that 

\begin{figure}
    \centering
    \includegraphics[width=\columnwidth]{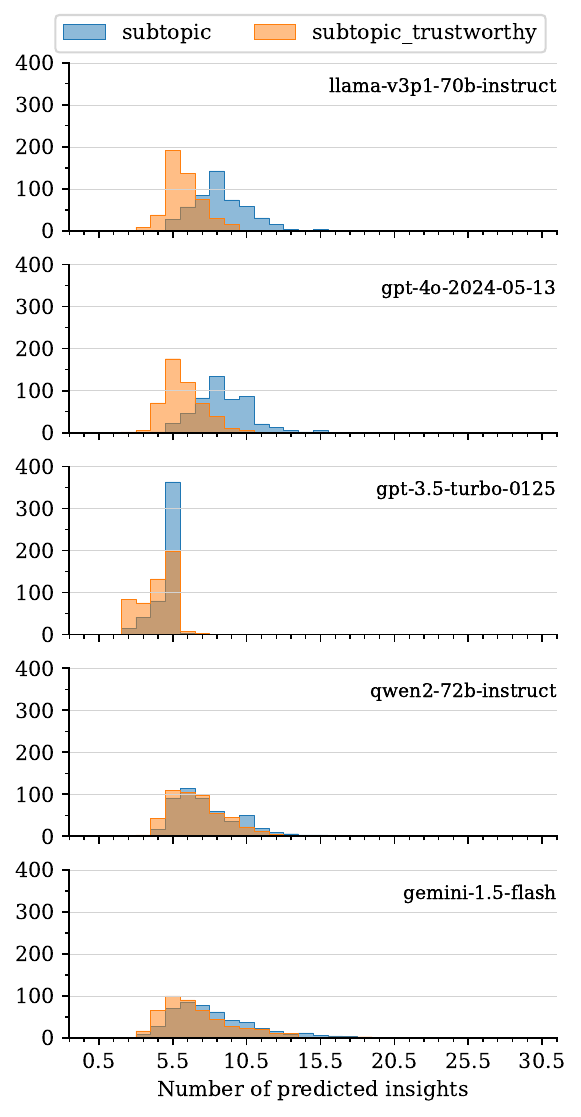}
    \caption{\textbf{Distribution of the number of predicted insights per model in the news domain for combinations of 5 documents (N=5)}.}
    \label{fig:app:dist-output-news}
    \vspace{-1em}
\end{figure}
\begin{figure}
    \centering
    \includegraphics[width=\columnwidth]{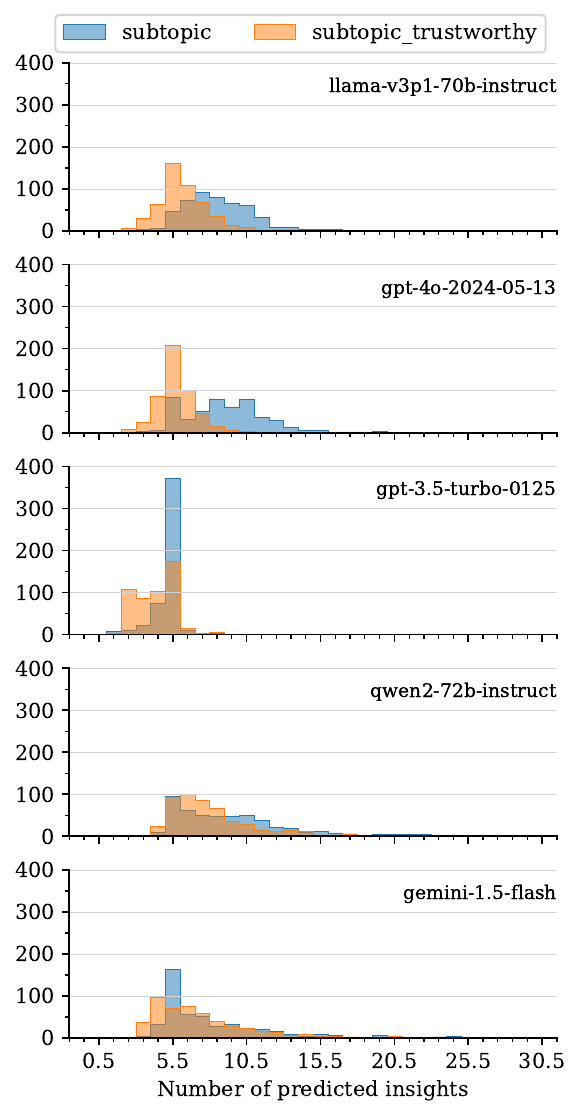}
    \caption{\textbf{Distribution of the number of predicted insights per model in the conv domain for combinations of 5 documents (N=5)}.}
    \label{fig:app:dist-output-conv}
\end{figure}

\subsection{F1-score Results}
\label{app:ssec:additional-results:f1score}

\begin{figure*}[tb]
    \centering
    \begin{subfigure}[b]{\columnwidth}
         \centering
         \includegraphics[width=\columnwidth]{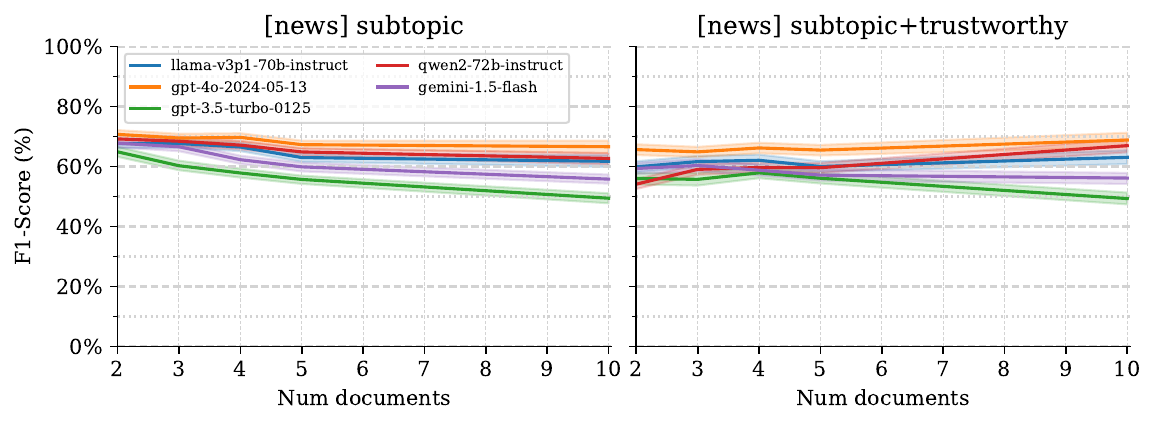}
        \caption{news}
        \label{fig:app:subtopic:f1-score:news}
    \end{subfigure}
    \begin{subfigure}[b]{\columnwidth}
        \centering
        \includegraphics[width=\columnwidth]{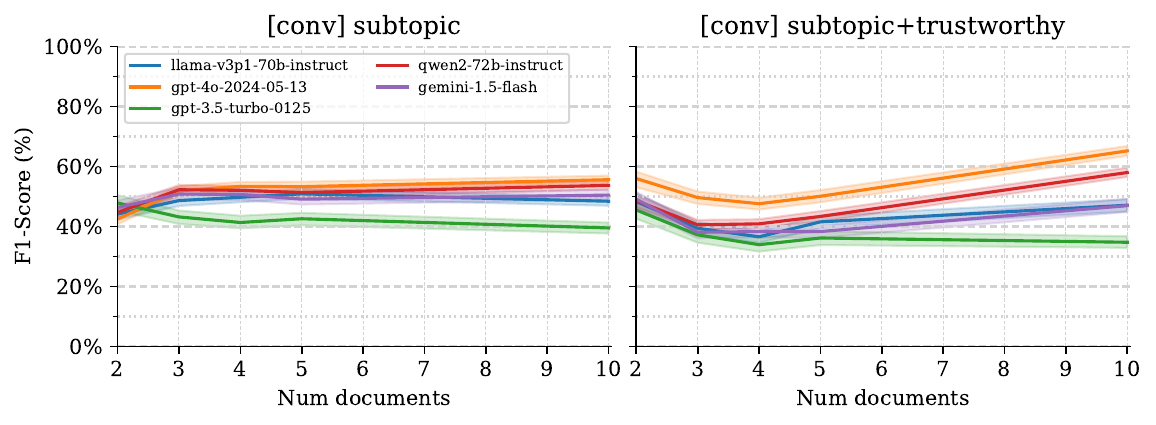}
        \caption{conv}
        \label{fig:app:subtopic:f1-score:conv}
    \end{subfigure}
    \caption{\textbf{F1-score as a function of the number of input documents }. Each line represents the mean value, with shaded areas indicating the 95\% confidence intervals.
    Overall, \fullchatgpt and \fullgemini are among the worst performers in both domains, due to their tendencies to generate overly short and long summaries, respectively. 
    Surprisingly, \fullqwen reveals to be on par with \fullgptfour, with the former exhibiting slightly superior performance overall (<5\% points).
    }
    \label{fig:app:subtopic:f1-score}
    \vspace{-1em}
\end{figure*}

Figure \ref{fig:app:subtopic:f1-score} reports the F1-scores obtained for both \subtopic and \subtopictrust settings across the news (Figure \ref{fig:app:subtopic:f1-score:news}) and conversation domain (Figure \ref{fig:app:subtopic:f1-score:conv}).

\subsection{Analysis Breakdown}
\label{app:ssec:analysis-preds-breakdown}

In this section, we report additional results about the types of errors. 
Tables \ref{tab:metric-bidirectional:pred-breakdown-conv-sub} summarize the percentage of correctness when using fc+pc label type. 
\input{appendix/tables/prediction_statistics/metric_bidirectional__preds-breakdown-conv-subtopic}

\subsection{Correctness vs Outputs}
\label{app:sec:data-output}

\begin{figure}[tb]
    \centering
    \begin{subfigure}[b]{\columnwidth}
        \includegraphics[width=\columnwidth]{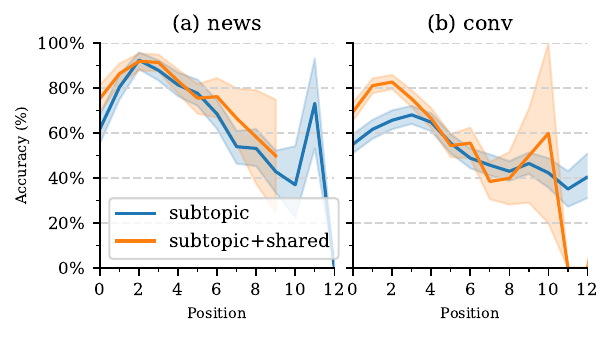}
        \postspace
        \minipostspace
        \label{app:fig:data-output:10docs:all-models:gptfo}
        \caption{\shortgptfour}
    \end{subfigure}
    \begin{subfigure}[b]{\columnwidth}
         \includegraphics[width=\columnwidth]{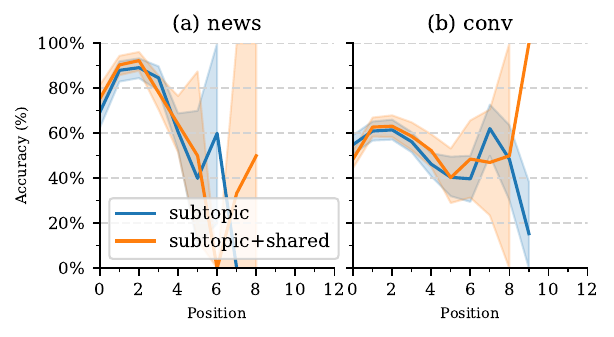}
        \postspace
        \minipostspace
        \caption{\shortchatgpt}
        \label{app:fig:data-output:10docs:all-models:chatgpt}
        
    \end{subfigure}
    \begin{subfigure}[b]{\columnwidth}
         \includegraphics[width=\columnwidth]{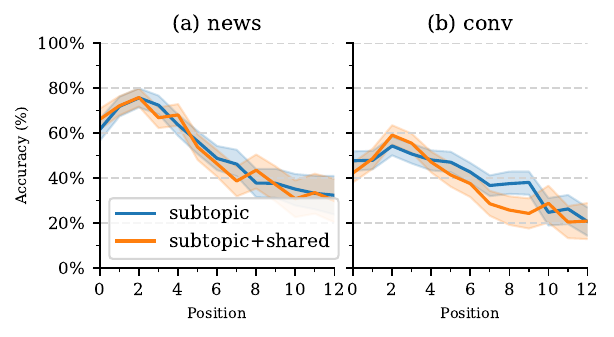}
        \postspace
        \minipostspace
         \caption{\shortgemini}
        \label{app:fig:data-output:10docs:all-models:gemini}
    \end{subfigure}
    \begin{subfigure}[b]{\columnwidth}
         \includegraphics[width=\columnwidth]{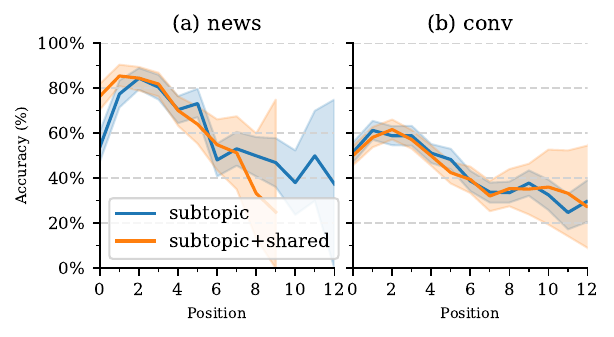}
        \postspace
        \minipostspace
          \caption{\shortllama}
        \label{app:fig:data-output:10docs:all-models:llama}
    \end{subfigure}
    \begin{subfigure}[b]{\columnwidth}
        \includegraphics[width=\columnwidth]{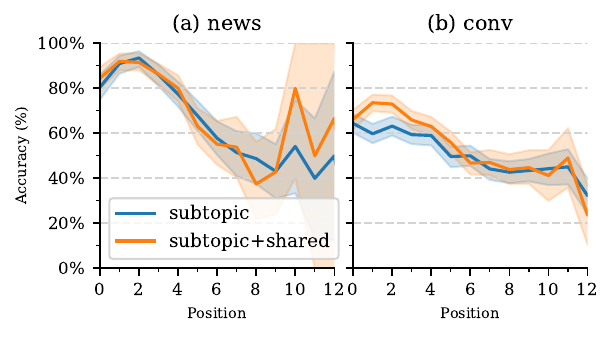}
        \postspace
        \minipostspace
        \caption{\shortqwen}
        \label{app:fig:data-output:10docs:all-models:qwen}
    \end{subfigure}
    \caption{\textbf{Accuracy rate of \llm-generated insights by position (when summarizing 10 input documents)}. Each solid line represents the mean value, with shaded areas indicating the 95\% confidence intervals. Overall, we observe the same pattern across all models: accuracy decreases as \llms generate more insights.}
    \label{app:fig:data-output:10docs:all-models}
\end{figure}

\begin{figure}[tb]
    \centering
    \begin{subfigure}[b]{\columnwidth}
        \centering
        \includegraphics[width=\columnwidth]{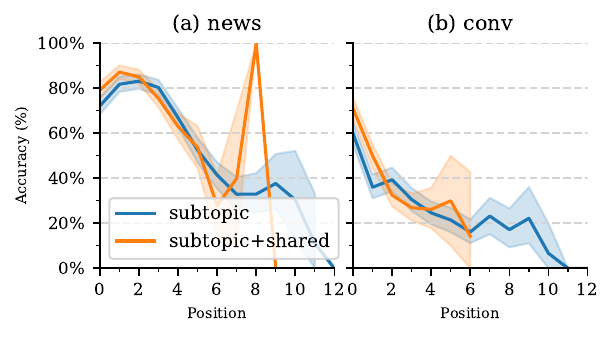}
        \postspace
        \minipostspace
        \caption{\shortgptfour}
        \label{app:fig:data-output:2docs:all-models:gptfo}
    \end{subfigure}
    \begin{subfigure}[b]{\columnwidth}
        \centering
         \includegraphics[width=\columnwidth]{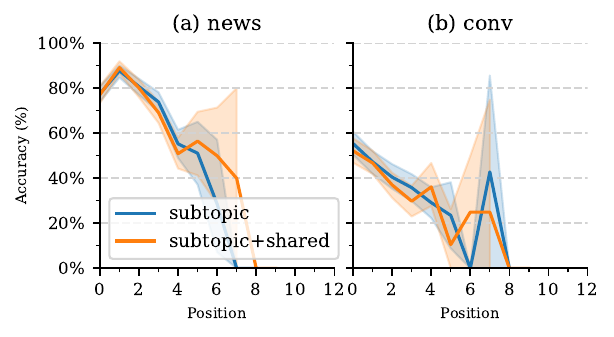}
        \postspace
        \minipostspace
         \caption{\shortchatgpt}
        \label{app:fig:data-output:2docs:all-models:chatgpt}
    \end{subfigure}
    \begin{subfigure}[b]{\columnwidth}
        \centering
         \includegraphics[width=\columnwidth]{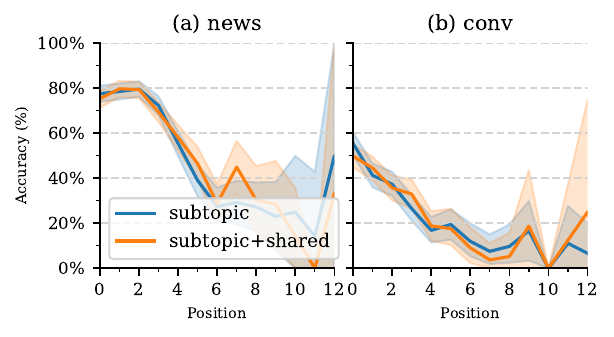}
        \postspace
        \minipostspace
         \caption{\shortgemini}
        \label{app:fig:data-output:2docs:all-models:gemini}
    \end{subfigure}
    \begin{subfigure}[b]{\columnwidth}
        \centering
         \includegraphics[width=\columnwidth]{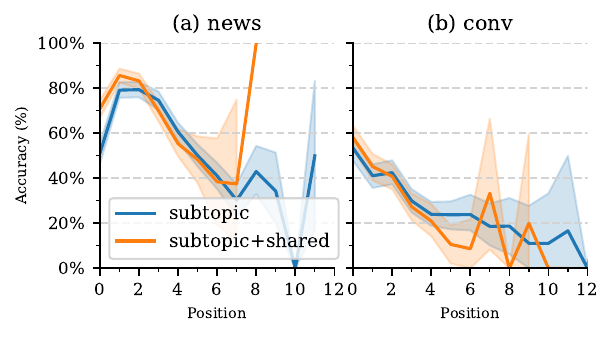}
        \postspace
        \minipostspace
         \caption{\shortllama}
        \label{app:fig:data-output:2docs:all-models:llama}
    \end{subfigure}
    \begin{subfigure}[b]{\columnwidth}
        \centering
        \includegraphics[width=\columnwidth]{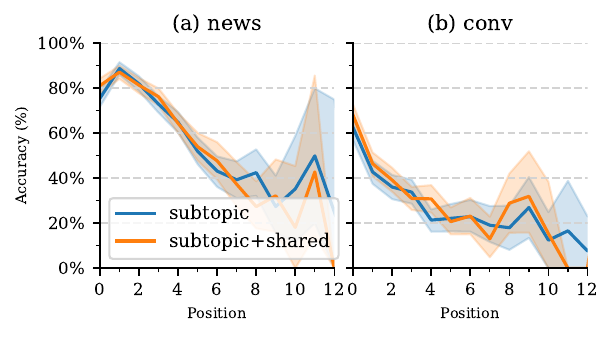}
        \postspace
        \minipostspace
        \caption{\shortqwen}
        \label{app:fig:data-output:2docs:all-models:qwen}
    \end{subfigure}
    \caption{\textbf{Accuracy rate of \llm-generated insights by position (when summarizing 2 input documents)}. Overall, we observe the same pattern across all models: accuracy decreases as \llms generate more insights.}
    \label{app:fig:data-output:2docs:all-models}
\end{figure}

In the main paper, we report the results with respect to a single model and when processing combination of 10 documents. 
In this section, we show evidence of the same behavior across all evaluated models (see Figure \ref{app:fig:data-output:10docs:all-models}), as well as for combination size N=2 (see Figure \ref{app:fig:data-output:2docs:all-models}).


\section{Hallucination Taxonomy's Annotation Details}
\label{app:sec:taxonomy}

To create the hallucination taxonomy, we collected human annotations for over 150 \llm-generated summaries and developed a taxonomy based on recurring mistakes. Due to the challenge of analyzing long documents, as noted in prior work~\citep{chang2024booookscoresystematicexplorationbooklength}, we limited our analysis to N=2, leaving the exploration of larger combinations for future research.

Two authors independently conducted the main analysis, manually inspecting over 700 predicted insights, which were nearly uniformly distributed across models, task focus, and domains. Each annotation sample included two input documents, the query containing the target subtopic, all reference insights, and the predicted insight. 
Annotators first determined whether the predicted coverage label (full, partial, or not covered) of the insight was correct and, then, provided a description for the type of error in the predicted insight. 
After gathering error descriptions, the annotators together discussed the observed patterns, defined the hallucination taxonomy based on those patterns, and categorized the hallucinated insights under each type.

\section{Hallucination Mitigation}
\label{app:sec:mitigation:experiments}

Based on the observed patterns during our manual annotation, we seek out to investigate the effectiveness of simple \textit{post-hoc} heuristics.
In particular, we make two observations: 
(1) with the exception of \shortchatgpt, \llms tend to over-generate insights (more than reference insights---See Figure \ref{fig:app:ratio-insights}) 
and (2) a large fraction of \llm mistakes is due to the generation of overly generic and/or repeated information. 
To assess the extent to which simple heuristics ---both \llm-based and rule-based---may help reducing such errors, we conduct a small-scale experiment and measure the impact in the measured recall and hallucination rate. 
Ideally, one would like mitigation approaches to reduce the hallucination rate with minimal impact on recall.

\subsection{Methods}
\label{app:ssec:mitigation:methods}
As previously mentioned, we propose two categories of methods: rule-based and model-based. 
The \textbf{rule-based methods} are faster to run and do not depend on existing \llms. 
The proposed rule-based methods are listed below:
\begin{itemize}
\item \textbf{Truncate summaries (Top-K)}: this method is based on our findings that \llms' insights generated earlier in the summary are more likely to be accurate than those generated later (see \ref{ssec:investigate:error-vs-output}). 
Given a summary composed of $I$ insights (expressed as bullet-points), this method truncates the summary, keeping the first $K$ insights.\footnote{
Early on, we also experimented with the removal of the first $L$ insights but found it to cause significant drops in recall and, therefore, chose not to include those results in the experiments.}
\end{itemize}

In addition to rule-based methods, we also explore three different \textbf{model-based methods}. 
As the name indicates, this class of methods rely on \llms and, hence, may be considered more time-consuming and costly than rule-based methods. 
However, these methods are also more versatile and nuanced facilitating the manipulation of semantic relationships between two texts. 
We explore these capabilities to mitigate some of the patterns observed in the \llms generations, such as generating insights that are unrelated to the subtopic, redundant, and/or paraphrases of the subtopic:
\begin{itemize}
    \item \textbf{Unrelated Subtopic (\texttt{st-unrelated})}: given the queried subtopic $q$ and a list of predicted insights (\ie a summary in bullet-point format), this method filters out the insights that are not related to the subtopic $q$. Particularly, for each predicted insight $i$, we zero-shot ask \fullgptmini whether $i$ is related to the subtopic $q$ and use the greedy decoded answer (yes/no) to determine whether to keep the insight or not. 
    Alternatively, we also experiment using different confidence thresholds $\alpha_u \in [0, 1]$, where we filter out the predicted insight if the likelihood assigned to the answer ``yes'' is below a threshold $\alpha_u$. We denote this method using the notation \texttt{st-unrelated}-$\alpha_u$.
    \item \textbf{Paraphrases of Subtopic (\texttt{st-paraphrase})}: the goal of this method is to identify and filter out the insights that have a similar meaning to subtopic $q$ or are superficial in nature. 
    Following previous work~\citep{michail2024paraphrasuscomprehensivebenchmark,Farquhar2024}, we explore the use of two different approaches to detect meaning similarity: (1) a zero-shot \llm-based approach (again instantiated using \fullgptmini) and (2) an entailment approach (instantiated using \texttt{microsoft/deberta-v2-xlarge-mnli}~\citep{he2021deberta}), which we denote \texttt{st-paraphrase} and \texttt{st-paraphrase-nli}, respectively.\footnote{Results reported in the main paper refer to the \llm approach.}
    By default, we consider the greedy answer or label when determining whether an insight $i$ is a paraphrase of subtopic $q$. 
    We also experiment using a the model's confidence $\alpha_p \in [0, 1]$ to regulate this prediction, which we denote  \texttt{st-paraphrase-}$\alpha_p$ or \texttt{st-paraphrase-nli-}$\alpha_p$ depending on whether we use the zero-shot or the entailment approach.
    \item \textbf{Redundant Bullet-points (\texttt{redundant})}: 
    To detect redundancy in the summaries, we re-use the previous zero-shot and entailment approaches but instead of using it to compare a subtopic with an insight, we use it to compare every pair of predicted insights in the summary. Whenever we find a redundant insight, we drop one of them. Like before, depending on whether the zero-shot \llm approach is being used or the nli approach, we refer to these methods as \texttt{redundant} and \texttt{redundant-nli}, appending a suffix -$\alpha_r$ to each method depending on the confidence threshold used. 
\end{itemize}

\subsection{Results}
\label{app:ssec:mitigation:results}

In this section, we present the full results corresponding to the hallucination mitigation. 
Table \ref{tab:app:mitigation:f1score:all-models} shows the average F1-score improvements after applying the various methods, whereas Tables \ref{tab:app:mitigation:recall:all-models} and \ref{tab:app:mitigation:err-rate:all-models} summarize the average improvements in terms of recall and hallucination rate, respectively.

\begin{table*}[tbh]
    \vspace{-1em}
    \small
    \centering
    \caption{\textbf{Absolute difference (in percentage points) in average F1-score after applying four simple mitigation methods to summaries generated from two input documents (N=2)}. 
    ``Top-k'' is the only rule-based method and all other methods are model-based. By default, model-based methods use greedily decoded with no confidence threshold (unless explicitly identified through ``-$\alpha$'' for $\alpha \in [0, 1]$. Models with ``-nli'' suffix are implemented using bidirectional entailment model (\texttt{microsoft/deberta-v2-xlarge-mnli}) opposed to general purpose \llm (\fullgptmini). All mitigation methods have little to no impact in terms of average F1-score ($\pm$3\%), highlighting the need for further research to address hallucinations in multi-document scenarios more systematically.}
    \label{tab:app:mitigation:f1score:all-models}
    \begin{tabular}{ll ccccc}
        \toprule
                                                        & \textbf{Strategy} &\textbf{\shortgemini} & \textbf{\shortchatgpt} & \textbf{\shortgptfour} & \textbf{\shortllama} & \textbf{\shortqwen} \\
        \midrule
        \multirow{7}{*}{\rotatebox{90}{news}}           & \texttt{top-5}            &  0.61\% &  0.09\% &  2.51\% &  0.42\% &  1.69\% \\ 
                                                        & \texttt{st-unrelated}     &  0.15\% & -0.37\% & -0.64\% & -0.27\% & -0.02\% \\
                                                        & \texttt{st-paraphrase}    & -1.69\% & -2.00\% & -1.19\% & -0.86\% & -1.19\% \\
                                                        & \texttt{redundant}        & -0.46\% & -0.02\% & -0.09\% & -0.41\% &  0.16\% \\
        \addlinespace
                                                        & \texttt{st-unrelated-0.8}       & -1.88\% &  -0.38\% & -2.61\% & -1.95\% & -1.49\% \\  
                                                        & \texttt{st-paraphrase-nli-0.6}  & 0.03\% & -0.01\% & 0.05\% & 0.09\% & 0.01\% \\
                                                        & \texttt{redundant-nli-0.6}      & -0.80\% & -1.23\% & -0.49\% & -0.98\% & -0.28\% \\
                                                        
        \midrule
        \multirow{7}{*}{\rotatebox{90}{conv}}           & \texttt{top-5}            &  1.37\% & 0.19\% & 2.28\% & 1.52\% & 1.52\% \\
                                                        & \texttt{st-unrelated}     &  0.24\% & 0.50\% & 0.76\% & 0.36\% & 0.63\% \\
                                                        & \texttt{st-paraphrase}    & -0.23\% & -0.77\% & -0.11\% & -0.29\% & -0.46\% \\  
                                                        & \texttt{redundant}        &  0.33\% & 0.11\% & 0.48\% & 0.40\% & 0.22\% \\                  
        \addlinespace
                                                        & \texttt{st-unrelated-0.8}             & 0.09\% & 0.64\% & 0.85\% & 0.43\% & 0.85\% \\
                                                        & \texttt{st-paraphrase-nli-0.6}        & 0.00\% & -0.04\% & 0.01\% & 0.00\% & 0.00\% \\
                                                        & \texttt{redundant-nli-0.6}            & 0.36\% & 0.53\% & 0.46\% & 0.18\% & -0.08\% \\                   
        \bottomrule
    \end{tabular}
\postspace
\minipostspace
\end{table*}
\begin{table*}[tbh]
    \vspace{-1em}
    \small
    \centering
    \caption{\textbf{Absolute difference (in percentage points) in average Recall after applying four simple mitigation methods to summaries generated from two input documents (N=2)}. 
    ``Top-k'' is the only rule-based method and all other methods are model-based. By default, model-based methods use greedily decoded with no confidence threshold (unless explicitly identified through ``-$\alpha$'' for $\alpha \in [0, 1]$. Models with ``-nli'' suffix are implemented using bidirectional entailment model (\texttt{microsoft/deberta-v2-xlarge-mnli}) opposed to general purpose \llm (\fullgptmini). Overall, we find slight drops in average recall (<10\%), we observe lower improvements (< 5\%) in average hallucination rate (in Table \ref{tab:app:mitigation:err-rate:all-models}), which highlights the complexity of mitigating these hallucination errors and the need for further exploration.} 
    \label{tab:app:mitigation:recall:all-models}
    \begin{tabular}{ll ccccc}
        \toprule
                                                        & \textbf{Strategy} &\textbf{\shortgemini} & \textbf{\shortchatgpt} & \textbf{\shortgptfour} & \textbf{\shortllama} & \textbf{\shortqwen} \\
        \midrule
        \multirow{7}{*}{\rotatebox{90}{news}}           & \texttt{top-5}            &  -2.72\% & -0.07\% & -4.66\% & -5.67\% & -3.58\% \\
                                                        & \texttt{st-unrelated}     &  -2.27\% & -2.82\% & -3.90\% & -2.58\% & -3.20\% \\       
                                                        & \texttt{st-paraphrase}    &  -3.30\% & -3.34\% & -3.14\% & -3.18\% & -3.10\% \\       
                                                        & \texttt{redundant}        &  -1.28\% & -0.12\% & -0.49\% & -1.96\% & -0.39\% \\       
        \addlinespace
                                                        & \texttt{st-unrelated-0.8}       & -5.11\% & -3.02\% & -8.55\% & -5.08\% & -7.40\% \\  
                                                        & \texttt{st-paraphrase-nli-0.6}  &  -0.05\% & -0.07\% & -0.03\% & 0.00\% & 0.00\% \\   
                                                        & \texttt{redundant-nli-0.6}      & -2.89\% & -1.94\% & -2.21\% & -4.12\% & -1.96\% \\  
        \midrule
        \multirow{7}{*}{\rotatebox{90}{conv}}           & \texttt{top-5}            &  -3.08\% & -0.15\% & -4.55\% & -2.64\% & -5.57\% \\
                                                        & \texttt{st-unrelated}     &  -0.15\% & -0.44\% & -0.29\% & -0.29\% & -0.15\% \\
                                                        & \texttt{st-paraphrase}    &  -2.05\% & -1.61\% & -0.44\% & -2.35\% & -1.47\% \\
                                                        & \texttt{redundant}        &  -1.32\% & 0.00\% & -0.88\% & -0.88\% & -1.03\% \\
   
        \addlinespace
                                                        & \texttt{st-unrelated-0.8}           & -0.59\% & -0.44\% & -0.44\% & -0.44\% & -0.15\% \\
                                                        & \texttt{st-paraphrase-nli-0.6}        & 0.00\% & -0.15\% & 0.00\% & 0.00\% & 0.00\% \\
                                                        & \texttt{redundant-nli-0.6}            & -3.96\% & -1.61\% & -3.23\% & -4.69\% & -4.25\% \\
        \bottomrule
    \end{tabular}
\postspace
\minipostspace
\end{table*}

\begin{table*}[tbh]
    \small
    \centering
    \caption{
    \textbf{Absolute difference (in percentage points) in average hallucination rate after applying four simple mitigation methods to summaries generated from two input documents (N=2)}. 
    ``Top-k'' is the only rule-based method and all other methods are model-based. By default, model-based methods use greedily decoded with no confidence threshold (unless explicitly identified through ``-$\alpha$'' for $\alpha \in [0, 1]$. Models with ``-nli'' suffix are implemented using bidirectional entailment model (\texttt{microsoft/deberta-v2-xlarge-mnli}) opposed to general purpose \llm (\fullgptmini). Overall, we find slight drops in average recall (< 10\%) (in Table \ref{tab:app:mitigation:recall:all-models}), we observe lower improvements (< 5\%) in average hallucination rate, which highlights the complexity of mitigating these hallucination errors and the need for further exploration.}
    \label{tab:app:mitigation:err-rate:all-models}
    \begin{tabular}{ll ccccc}
        \toprule
                        & \textbf{Strategy} &\textbf{\shortgemini} & \textbf{\shortchatgpt} & \textbf{\shortgptfour} & \textbf{\shortllama} & \textbf{\shortqwen} \\
        \midrule
        \multirow{7}{*}{\rotatebox{90}{news}}           & \texttt{top-5}            & 2.29\% & 0.18\% & 6.09\% & 3.58\% & 4.28\% \\
                                                        & \texttt{st-unrelated}     & 0.61\% & 0.93\% & 0.59\% & 0.43\% & 0.98\% \\
                                                        & \texttt{st-paraphrase}    & 0.67\% & 0.87\% & 0.28\% & 0.92\% & 0.44\% \\
                                                        & \texttt{redundant}        & 0.14\% & 0.19\% & 0.17\% & 0.66\% & 0.53\% \\
        \addlinespace
                                                        & \texttt{st-unrelated-0.8}       &  0.70\% & 1.17\% & 1.87\% & 0.49\% & 2.98\% \\
                                                        & \texttt{st-paraphrase-nli-0.6}  &  0.14\% & 0.14\% & 0.11\% & 0.17\% & 0.01\% \\
                                                        & \texttt{redundant-nli}          &  1.10\% & 0.23\% & 0.57\% & 1.26\% & 0.92\% \\
                                                        
        \midrule
        \multirow{7}{*}{\rotatebox{90}{conv}}           & \texttt{top-5}            &  1.26\% & 0.17\% & 2.21\% & 1.44\% & 1.61\% \\
                                                        & \texttt{st-unrelated}     &  0.22\% & 0.66\% & 0.80\% & 0.42\% & 0.61\% \\
                                                        & \texttt{st-paraphrase}    & 0.80\% & -0.61\% & -0.04\% & 0.16\% & -0.24\% \\
                                                        & \texttt{redundant}        & 0.70\% & 0.12\% & 0.65\% & 0.53\% & 0.40\% \\
        \addlinespace
                                                        & \texttt{st-unrelated-0.8}       & 0.25\% & 0.84\% & 0.89\% & 0.52\% & 0.83\% \\
                                                        & \texttt{st-paraphrase-nli-0.6}  & -0.00\% & -0.00\% & 0.01\% & -0.00\% & -0.00\% \\
                                                        & \texttt{redundant-nli}          & 0.95\% & 1.12\% & 0.83\% & 0.86\% & 0.52\% \\            
        \bottomrule
    \end{tabular}
\postspace
\minipostspace
\end{table*}

\end{document}

%% file: main/tables/hallucination-annotations-news-subtopic_and_trust.tex
\begin{table}[tb]
    \small
    \centering
    \caption{\textbf{Observed hallucinations (and their frequency) when summarizing news articles for different task focus (\subtopic and \subtopictrust)}. Reported values represent the fraction of error mistakes that are categorized as either \textbf{P}edantic, \textbf{I}nstruction Inconsistency, \textbf{C}ontext Inconsistency, and \textbf{F}abrication. We also report the \textbf{T}otal number of errors for each \llm.}
    \label{tab:main:annotations:subtopic-and-trust:news}
    \begin{tabular}{clcccc c}
        \toprule
        & \textbf{Model} & \textbf{P} & \textbf{I} & \textbf{C} & \textbf{F} & \textbf{T}\\
        \midrule
        \multirow{5}{*}{\rotatebox{90}{subtopic}} &
          \texttt{GPT-3.5} & 51.51 & 38.71 & 19.35 & 0.00 & 31 \\
        & \shortgptfour & 61.88 & 52.73 & 21.81 & 0.00 & 55 \\
        & \texttt{Gemini}  & 78.72 & 23.40 & 10.64 & 8.51 & 47 \\
        & \texttt{Llama 3.1}  & 60.00 & 72.00 & 16.00 & 4.00 & 25 \\
        & \texttt{Qwen 2}    & 53.33 & 86.67 & 36.37 & 3.33 & 60 \\
        \midrule
        \multirow{5}{*}{\rotatebox{90}{shared}} &
          \texttt{GPT-3.5}  & 38.46 & 75.00 & 12.82 & 0.00 & 32 \\
        & \shortgptfour  & 28.15 & 79.49 &  9.30 & 0.00 & 39 \\
        & \texttt{Gemini}  & 52.73 & 70.91 &  9.09 & 0.00 & 55 \\
        & \texttt{Llama 3.1}  & 29.27 & 70.73 & 17.07 & 2.4 & 41 \\
        & \texttt{Qwen 2}  & 15.51 & 86.20 & 17.24 & 1.7 & 58 \\
        \bottomrule
    \end{tabular}
\end{table}

%% file: appendix/tables/topics_summhay.tex
\begin{table*}[tb]
\centering
\begin{tabular}{p{0.45\linewidth} p{0.45\linewidth}}
\toprule
\textbf{\newsdataset} & \textbf{\convdataset} \\
\midrule
Comprehensive Analysis of Dodge V8 Muscle Cars: Evolution, Performance, and Future Trends & A 25-minute meeting between a doctor and a patient discussing the patient's recent health concerns and treatment options \\
\addlinespace
Economic and Regulatory Implications of the Silicon Valley Bank Collapse & A 20-minute meeting between a project manager, two team members, and a stakeholder discussing the progress of a software development project at IBM \\
\addlinespace
Strategic Growth and Operational Changes at Foot Locker & A 50-minute debate between three colleagues that work at a big company discussing the pros and cons of remote work \\
\addlinespace
Recent Developments and Challenges in Twitter's Algorithms and Services & A 30-minute phone conversation between a Salesforce sales representative and either a current, prospective, or former customer \\
\addlinespace
Current Financial Market Dynamics and Institutional Responses & A 25-minute study group session where three students discuss their strategies and insights for an upcoming exam \\
\bottomrule
\end{tabular}
\caption{\textbf{List of topics included in the SummHay dataset~\citep{summhay--laban-et-al-2024}, highlighting the thematic scope and domain coverage}. The topics proposed in the news domain are entity-centric, grounded on well-known entities (\eg Dodge V8 Muscle cars, Silicon Valley Bank, Twitter's, Foot Locker), but also more likely to be quantitative. Conversely, the topics in the conversation domain include 2-4 participants and are more broad.}
\label{tab:app:topics-summhay}
\end{table*}

%% file: appendix/examples/ref-insights.tex
\subsubsection{What is an insight?}  
\label{app:sssec:what-is-an-insight}

Next, we enumerate a few examples of insight-level annotations found in the original dataset.

\begin{itemize}
    \item \textbf{News domain}: The insights in this domain are generally more quantitative, richer in details, and descriptive of the events in the documents. 
    However, we notice that certain subtopics tend to be more abstract and subjective (\eg insights related to historical context or NHRA regulations). 
    Moreover, despite the insight quality checks conducted in the original dataset, we find evidence that insights are not always independent and sometimes exhibit large lexical and even semantic overlap (as emphasized by the reference insights related to \textit{Foot Locker}). 
    Specifically, based on our manual analysis, we notice that in examples containing partially overlapping insights, \llms tended to generate a single insight (instead of multiple) and therefore naturally exhibited lower recall for that summary.
        \begin{enumerate}
            \itemsep-0.3em
            \item {\small``Dodge’s switch to E85 for the Demon 170 marks a significant shift from their historical reliance on gasoline, driven by the high performance demands of the vehicle.''}
            \item {\small ``The journey from the original 426 Hemi in 1964 to the 2023 Demon 170 highlights significant engineering milestones, such as the introduction of the supercharged Hellcat in 2015, producing 707 horsepower.''}
            \item {\small ``Many tech startups suddenly faced cash flow issues as they scrambled to find new banking partners after SVB failed. Some had difficulty understanding new banking systems and integrating them with their accounting software, causing delays in operations.''}
            \item {\small ``Foot Locker reported Q4 2022 sales of \$2.33 billion, exceeding consensus expectations of \$2.15 billion, despite a 0.3\% year-over-year decline.''}
            \item {\small ``Dillon's leadership is set to position Foot Locker for growth in 2024 and beyond, focusing on expanding wallet share and broadening customer reach.''}
            \item {\small ``Despite Tesla's reputation, the expanding EV market has allowed competitors to gain traction, diluting Tesla's early advantage.''}
            \item {\small ``During the 2021 forest fires in Turkiye, the '\#HelpTurkey' hashtag on Twitter generated a nationalistic reaction as it was perceived to imply that the country could not handle the crisis, highlighting the power of hashtags in influencing public perception.''}
            \item {\small ``Banks are expected to adopt more diversified financial models to mitigate future risks. For example, banks might integrate asset-backed lending and communal investment pools to spread out risk effectively.''}
            \item {\small ``Foot Locker plans to ensure that approximately 40\% of its revenue will come from non-Nike brands by 2026 to reduce dependency on any single brand.''}
            \item {\small ``Foot Locker is focusing on higher-margin non-Nike products to improve profitability while reducing reliance on any single footwear brand.''}
            \item {\small ``Foot Locker expects to double its revenue from non-Nike brands by focusing on sneakers that cater to different occasions and activities, thereby appealing to a wider audience.''}
            \item {\small ``The company noted a broadening consumer base that requires a wider range of brands, including non-Nike offerings, to meet various footwear needs.''}
            \item{\small ``The company plans to broaden its assortment of lifestyle and performance shoes from various brands during the holiday seasons to cater to diverse consumer preferences.''}
        \end{enumerate}

    \item \textbf{Conversation domain}: Unlike the news domain, the conversation (conv) dataset tend to describe multi-turn interactions between participants in ordinary situations (\eg doctor appointments, company meetings, student debates). 
    Notably, while the reference insights remain detail-specific, aiming to accurately capture the participants' interactions and responses (\eg by specifying time, duration, cost, or other specific details related to the interactions), they are participant-agnostic. That is, they are not tailored to a participant with a specific name. 
    Additionally, note how some insights simultaneously convey multiple pieces of information, rather than focusing on a single unit of information. 
        \begin{enumerate}
            \itemsep-0.3em
            \item {\small ``The doctor questions whether the symptoms have been getting worse, better, or staying the same, and the patient indicates that they have been gradually worsening over the last few days.''}
            \item {\small ``The doctor asks the patient to rate the severity of their symptoms on a scale from 1 to 10, and the patient rates their symptoms as a 7.''}
            \item {\small ``The project manager notes that the user manual and API documentation are 50\% complete, with a target to finish by the end of the month.''}
            \item {\small ``The project manager allocates additional resources to a critical task that requires more attention, assigning 2 extra developers to ensure timely completion.''}
            \item {\small ``They explore the necessity of using task management tools like Asana or Trello to assign tasks, track progress, and hold team members accountable for their deliverables.''}
            \item {\small ``The sales rep asks about the primary sources of customer data for the business, and the customer lists their website forms, social media platforms, and email marketing campaigns as the main sources.''}
            \item {\small ``The sales representative and customer agree to schedule a live demonstration for the following Tuesday at 2 PM.''}
            \item {\small ``The sales rep inquires about the tools the customer is currently using to manage their business processes, and the customer responds that they are using a mix of Excel sheets and a simple CRM software from a lesser-known vendor.''}
            \item {\small ``One student offers to scan their handwritten notes and convert them into PDFs, which they believe might be useful for others who prefer reading handwritten material.''}
            \item {\small ``One student finds that their best study time is at night, typically starting their sessions at 10 PM and studying until 2 AM.''}
        \end{enumerate}
\end{itemize}

%% file: appendix/tables/dataset_statistics/news-domain.tex
\begin{table*}[tb]
\centering
\small
\caption{\textbf{Different statistics for varying number of document combinations (N) in the proposed evaluation benchmark for the news domain.} In addition to the total number of combinations (\texttt{\# combinations}), we report the average (and standard deviation) over the number of combinations for the total document length in tokens (\texttt{total length}), as well as for the number of insights and subtopics included in each combination.}
\label{tab:dataset-statistics-news}
\begin{tabular}{lccccc}
\toprule
    & N = 2 & N = 3 & N = 4 & N = 5 & N = 10 \\
\midrule
\# combinations             & 1.5k                  & 1.5k                  & 1.5k                  & 1.5k & 1.2k \\
total length                & $1766.2_{\pm 243.2}$  & $2626.9_{\pm 328.0}$  & $3522.8_{\pm 397.3}$  & $4363.6_{\pm 464.6}$ & $8643.7_{\pm 895.0}$ \\
\addlinespace
\# subtopics                & $2.9_{\pm 0.4}$       & $3.6_{\pm 0.5}$       & $4.3_{\pm 0.7}$       & $4.8_{\pm 0.7}$ & $6.8_{\pm 0.9}$ \\
\# shared subtopics         & $1.1_{\pm 0.4}$       & $1.4_{\pm 0.5}$       & $1.7_{\pm 0.6}$       & $2.0_{\pm 0.6}$ & $3.9_{\pm 0.7}$ \\
\addlinespace
\# insights                 & $11.4_{\pm 1.8}$      & $14.8_{\pm 2.2}$      & $18.4_{\pm 2.6}$      & $21.6_{\pm 3.0}$ & $34.5_{\pm 3.5}$ \\
\# subtopic insights        & $5.2_{\pm 0.8}$       & $5.6_{\pm 0.8}$       & $6.6_{\pm 1.0}$       & $7.3_{\pm 1.0}$ & $9.2_{\pm 0.8}$ \\
\# shared insights          & $2.8_{\pm 1.0}$       & $3.9_{\pm 1.4}$       & $4.9_{\pm 1.6}$       & $6.3_{\pm 1.8}$ & $14.4_{\pm 2.4}$ \\
\# subtopic shared insights &  $2.6_{\pm 0.6}$      & $2.4_{\pm 0.6}$       & $2.8_{\pm 0.9}$       & $3.5_{\pm 1.0}$ & $7.0_{\pm 1.0}$ \\
\bottomrule
\end{tabular}
\end{table*}

%% file: appendix/tables/dataset_statistics/conversation-domain.tex
\begin{table*}[tb]
\centering
\small
\caption{\textbf{Different statistics for varying number of document combinations (N) in the proposed evaluation benchmark for the conversation domain.} In addition to the total number of combinations (\texttt{\# combinations}), we report the average (and standard deviation) over the number of combinations for the total document length in tokens (\texttt{total length}), as well as for the number of insights and subtopics included in each combination.}
\label{tab:dataset-statistics-conversation}
\begin{tabular}{lccccc}
\toprule
            & N = 2 & N = 3 & N = 4 & N = 5 & N = 10 \\
\midrule
\# combinations         & 341                   & 1.5k                  & 1.5k                  & 1.5k                  & 1.5k \\
\addlinespace
total length  & $2054.6_{\pm 235.2}$  & $3017.5_{\pm 306.5}$  & $3992.3_{\pm 354.3}$  & $4947.4_{\pm 405.8}$  & $9870.2_{\pm 622.2}$ \\
\addlinespace
\# subtopics        & $2.9_{\pm 0.3}$       & $3.7_{\pm 0.5}$       & $4.4_{\pm 0.6}$       & $5.0_{\pm 0.7}$       & $7.3_{\pm 0.9}$ \\
\# shared subtopics & $1.1_{\pm 0.3}$       & $1.3_{\pm 0.5}$       & $1.6_{\pm 0.6}$       & $1.9_{\pm 0.6}$       & $3.8_{\pm 0.8}$ \\
\addlinespace
\# insights         & $5.2_{\pm 0.7}$       & $8.5_{\pm 0.9}$       & $11.2_{\pm 1.1}$      & $13.4_{\pm 1.2}$      & $21.7_{\pm 1.6}$ \\
\# subtopic insights& $2.0_{\pm 0.0}$       & $3.6_{\pm 0.5}$       & $4.5_{\pm 0.7}$       & $5.0_{\pm 0.7}$       & $5.9_{\pm 0.3}$ \\
\# shared insights  & $2.0_{\pm 0.2}$       & $2.2_{\pm 0.4}$       & $2.6_{\pm 0.7}$       & $3.3_{\pm 0.9}$       & $7.3_{\pm 1.3}$ \\
\# subtopic shared insights &  $2.0_{\pm 0.0}$ & $2.0_{\pm 0.2}$    & $2.3_{\pm 0.5}$       & $2.8_{\pm 0.7}$       & $5.3_{\pm 0.7}$ \\
\bottomrule
\end{tabular}
\end{table*}

%% file: appendix/prompts/summary_prompt.tex
\begin{figure*}[tb]
\centering
\begin{tcolorbox}[fonttitle=\fontfamily{pbk}\selectfont\bfseries,
                  fontupper=\fontfamily{ppl}\selectfont\itshape,
                  fontlower=\fontfamily{put}\selectfont\scshape,
                  title= News Summarization - \subtopic Prompt,
                  width=\linewidth,
                  arc=1mm, auto outer arc]
\begin{Verbatim}[breaklines=true, breaksymbol={}]
You are given {{n_articles}} news articles about the main subject "{{topic}}".

```
Article 1:
{{article_1}}

Article 2:
{{article_2}}
{{remaining_articles}}
```

Your objective is to clearly and concisely summarize all insights regarding the topic of "{{subtopic}}". If you don't find insights regarding the topic of "{{subtopic}}", return "No insights found".

Careful:
- [Format] You should format your summary as a bullet point list, where each bullet point is a different insight consisting of a single sentence. Represent each bullet point using "-". If you don't find related insights, write "No insights found" and nothing else.
- [Length] Your summary should be concise and clear.
\end{Verbatim}
\end{tcolorbox}
\caption{\textbf{Prompt used to summarize news articles}. The \subtopic prompt instructs the model to summarize \textit{any} insight in the documents that relates to the specified \texttt{\{\{subtopic\}\}}. \texttt{\{\{placeholders\}\}} will be replaced with corresponding values.}
\label{fig:prompt:summary-news:subtopic}
\vspace{-1em}
\end{figure*}
\begin{figure*}[tb]
\centering
\begin{tcolorbox}[fonttitle=\fontfamily{pbk}\selectfont\bfseries,
                  fontupper=\fontfamily{ppl}\selectfont\itshape,
                  fontlower=\fontfamily{put}\selectfont\scshape,
                  title=Conv Summarization - \subtopic Prompt,
                  width=\linewidth,
                  arc=1mm, auto outer arc]
\begin{Verbatim}[breaklines=true, breaksymbol={}]
You are given {{n_conversations}} conversations about the following scenario: "{{topic}}".
The conversations involve the following participants: {{participants}}. In each conversation, the participants might be different, with different names, but all the conversations fall into the same scenario.

```
Conversation 1:
{{conversation_1}}

Conversation 2:
{{conversation_2}}
{{remaining_conversations}}
```

Your objective is to clearly and concisely summarize all insights in the document regarding the topic of "{{subtopic}}".
If you don't find insights regarding the topic of "{{subtopic}}", write "No insights found".

Careful:
- [Format] You should format your summary as a bullet point list, where each bullet point is a different insight consisting of a single sentence. Represent each bullet point using "-". If you don't find related insights, write "No insights found" and nothing else.
- [Length] Your summary should be concise and clear.
\end{Verbatim}
\end{tcolorbox}
\caption{\textbf{Prompt used to summarize multiple conversations}. The \subtopic prompt instructs the model to summarize \textit{any} insight in the documents that relates to the specified \texttt{\{\{subtopic\}\}}. \texttt{\{\{placeholders\}\}} will be replaced with corresponding values.}
\label{fig:prompt:summary-conv:subtopic}
\vspace{-1em}
\end{figure*}
\begin{figure*}[tb]
\centering
\begin{tcolorbox}[fonttitle=\fontfamily{pbk}\selectfont\bfseries,
                  fontupper=\fontfamily{ppl}\selectfont\itshape,
                  fontlower=\fontfamily{put}\selectfont\scshape,
                  title=News Summarization - \subtopictrust Prompt,
                  width=\linewidth,
                  arc=1mm, auto outer arc]
\begin{Verbatim}[breaklines=true, breaksymbol={}]
You are given {{n_articles}} news articles about the main subject "{{topic}}".

```
Article 1:
{{article_1}}

Article 2:
{{article_2}}
{{remaining_articles}}
```

Your objective is to clearly and concisely summarize all insights regarding the topic of "{{subtopic}}" that are mentioned in at least two articles. In other words, your summary should include any insight that is related to the topic and is  mentioned in two or more articles.
If you don't find insights regarding the topic of "{{subtopic}}", return "No insights found".

Careful:
- [Format] You should format your summary as a bullet point list, where each bullet point is a different insight consisting of a single sentence. Represent each bullet point using "-". If you don't find related insights, write "No insights found" and nothing else.
- [Length] Your summary should be concise and clear.
\end{Verbatim}
\end{tcolorbox}
\caption{\textbf{Prompt used to summarize commonalities across news articles}. The \subtopictrust prompt instructs the model to summarize insights that are \textit{shared} across two or more documents and that relate to the specified \texttt{\{\{subtopic\}\}}. \texttt{\{\{placeholders\}\}} will be replaced with corresponding values.}
\label{fig:prompt:summary-news:subtopic+trustworthy}
\vspace{-1em}
\end{figure*}
\begin{figure*}[tb]
\centering
\begin{tcolorbox}[fonttitle=\fontfamily{pbk}\selectfont\bfseries,
                  fontupper=\fontfamily{ppl}\selectfont\itshape,
                  fontlower=\fontfamily{put}\selectfont\scshape,
                  title=Conv Summarization - \subtopictrust Prompt,
                  width=\linewidth,
                  arc=1mm, auto outer arc]
\begin{Verbatim}[breaklines=true, breaksymbol={}]
You are given {{n_articles}} conversations about the following scenario: "{{topic}}".
The conversations involve the following participants: {{participants}}. In each conversation, the participants might be different, with different names, but all the conversations fall into the same scenario.

```
Conversation 1:
{{conversation_1}}

Conversation 2:
{{conversation_2}}
{{remaining_conversations}}
```

Your objective is to clearly and concisely summarize all insights regarding the topic of "{{subtopic}}" that are mentioned in at least two conversations. In other words, your summary should include any insight that is related to the topic and is mentioned in two or more conversations.
If you don't find insights regarding the topic of "{{subtopic}}", write "No insights found".

Careful:
- [Format] You should format your summary as a bullet point list, where each bullet point is a different insight consisting of a single sentence. Represent each bullet point using "-". If you don't find related insights, write "No insights found" and nothing else.
- [Length] Your summary should be concise and clear.
\end{Verbatim}
\end{tcolorbox}
\caption{\textbf{Prompt used to summarize commonalities across conversations}. The \subtopictrust prompt instructs the model to summarize insights that are \textit{shared} across two or more documents and that relate to the specified \texttt{\{\{subtopic\}\}}. \texttt{\{\{placeholders\}\}} will be replaced with corresponding values.}
\label{fig:prompt:summary-conv:subtopic+trustworthy}
\vspace{-1em}
\end{figure*}

%% file: appendix/prompts/evaluation_prompt.tex
\begin{figure*}[tp]
\centering
\small
\begin{tcolorbox}[fonttitle=\fontfamily{pbk}\selectfont\bfseries,
                  fontupper=\fontsize{8}{9}\fontfamily{ppl}\selectfont\itshape,
                  fontlower=\fontfamily{put}\selectfont\scshape,
                  title=Evaluation Prompt,
                  width=\linewidth,
                  arc=1mm, auto outer arc]
\begin{Verbatim}[breaklines=true, breaksymbol={}]
You are given a list of bullet points (each with a unique number), and a specific reference insight. Your objective is to determine whether the reference insight is covered in any of the bullet points. You must further determine if the insight is partially covered ("PARTIAL_COVERAGE") or fully covered ("FULL_COVERAGE") by the bullet points. If the insight is not covered at all, you must return "NO_COVERAGE". See examples below:

Example Reference Insight 1: "The doctor asks the patient about their medical history".

Example Bullet Points 1:
{"bullets": [
    {"bullet_id": 1, "text": "The patient often mention that they are worried about medication side-effect."},
    {"bullet_id": 2, "text": "The doctor and patient spend time going over symptoms, particularly the initial symptoms and the progression in the last few months."},
    {"bullet_id": 3, "text": "The doctor and patient discuss medical history within the patient's family, with the patient often unaware that some of the conditions are hereditary."}
]}

Example Output 1:
{"coverage": "FULL_COVERAGE", "bullet_id": 3}

Example Reference Insight 2: "The doctor asks the patient about their medical history".

Example Bullet Points 2:
{"bullets": [
    {"bullet_id": 1, "text": "The patient often mention that they are worried about medication side-effect."},
    {"bullet_id": 2, "text": "The doctor and patient spend time going over symptoms, particularly the initial symptoms and the progression in the last few months."}
]}

Example Output 2:
{"coverage": "NO_COVERAGE", "bullet_id": "NA"}

Example Reference Insight 3: "The doctor asks the patient about their medical history".

Example Bullet Points 3:
{"bullets": [
    {"bullet_id": 1, "text": "The patient often mention that they are worried about medication side-effect."},
    {"bullet_id": 2, "text": "The doctor and patient catch up after a long time, with the patient mentioning feeling unwell for a while, and knowing of other family member's similar experiences."},
    {"bullet_id": 3, "text": "The doctor and patient spend time going over symptoms, particularly the initial symptoms and the progression in the last few months."}
]}

Example Output 3:
{"coverage": "PARTIAL_COVERAGE", "bullet_id": 2}

Now complete the task for the following insight and bullet points:

Reference Insight:
{{reference_insight}}

Bullet Points:
{{candidate_insights}}

Requirements:
- Do not hallucinate that the insight is covered by the bullet points if it is not.
- Your response should only be the JSON output in the format above, such that it can directly parsed by Python's json module. DO NOT OUTPUT ANY EXPLANATION OR ANYTHING THAT IS NOT THE JSON RESPONSE.
\end{Verbatim}
\end{tcolorbox}
\postspace
\minipostspace
\caption{\textbf{Few-shot prompt to automatically evaluate information coverage}. Proposed by~\citeauthor{summhay--laban-et-al-2024}, the prompt  determines whether the information conveyed by the \texttt{\{\{reference\_insight\}\}} is fully, partially, or not covered by any of the \texttt{\{\{candidate\_insights\}\}}.} 
\label{fig:prompt:eval-prompt}

\end{figure*}

%% file: appendix/tables/prediction_statistics/metric_bidirectional__preds-breakdown-conv-subtopic.tex
\begin{table*}
\caption{\textbf{Analysis of summarizer predictions in the conversation domain for varying number of in-context documents, when using the subtopic prompt}.}
\label{tab:metric-bidirectional:pred-breakdown-conv-sub}
\centering
\begin{tabular}{llllll}
\toprule
\textbf{Summarizer} & \textbf{\% Shared} & \textbf{\% Subtopic} & \textbf{\% Shared-Sub} & \textbf{\% Context}  & \textbf{\% Not Context} \\
\midrule
\textbf{N = 2} &&&&&\\
gpt-3.5-turbo-0125 & $37.85_{\pm 23.66}$ & $37.61_{\pm 23.56}$ & $37.61_{\pm 23.56}$ & $45.48_{\pm 23.91}$ & $54.52_{\pm 23.91}$ \\
gpt-4o-2024-05-13 & $29.43_{\pm 14.37}$ & $29.34_{\pm 14.29}$ & $29.34_{\pm 14.29}$ & $35.9_{\pm 17.18}$ & $64.1_{\pm 17.18}$ \\
llama-v3p1-70b & $32.01_{\pm 17.17}$ & $31.93_{\pm 17.13}$ & $31.93_{\pm 17.13}$ & $38.03_{\pm 18.53}$ & $61.97_{\pm 18.53}$ \\
\addlinespace

\textbf{N = 3} &&&&&\\
gpt-3.5-turbo-0125 & $28.48_{\pm 18.87}$ & $40.42_{\pm 25.18}$ & $28.0_{\pm 18.75}$ & $49.74_{\pm 26.02}$ & $50.26_{\pm 26.02}$ \\
gpt-4o-2024-05-13 & $25.37_{\pm 12.84}$ & $40.79_{\pm 17.86}$ & $25.17_{\pm 12.86}$ & $49.33_{\pm 20.53}$ & $50.67_{\pm 20.53}$ \\
llama-v3p1-70b & $25.24_{\pm 13.22}$ & $39.18_{\pm 18.64}$ & $24.92_{\pm 13.01}$ & $46.43_{\pm 20.78}$ & $53.57_{\pm 20.78}$ \\
\addlinespace

\textbf{N = 4} &&&&&\\
gpt-3.5-turbo-0125 & $29.76_{\pm 21.23}$ & $43.04_{\pm 26.12}$ & $28.77_{\pm 20.81}$ & $51.72_{\pm 26.44}$ & $48.28_{\pm 26.44}$ \\
gpt-4o-2024-05-13 & $26.31_{\pm 13.0}$ & $43.67_{\pm 17.23}$ & $25.65_{\pm 12.7}$ & $52.93_{\pm 18.78}$ & $47.07_{\pm 18.78}$ \\
llama-v3p1-70b & $25.56_{\pm 13.56}$ & $41.6_{\pm 18.27}$ & $24.85_{\pm 13.13}$ & $48.68_{\pm 19.65}$ & $51.32_{\pm 19.65}$ \\
\addlinespace

\textbf{N = 5} &&&&&\\
gpt-3.5-turbo-0125 & $32.75_{\pm 22.17}$ & $44.95_{\pm 25.24}$ & $31.84_{\pm 21.71}$ & $53.55_{\pm 26.63}$ & $46.45_{\pm 26.63}$ \\
gpt-4o-2024-05-13 & $28.41_{\pm 12.9}$ & $44.13_{\pm 16.78}$ & $27.9_{\pm 12.88}$ & $52.26_{\pm 18.91}$ & $47.74_{\pm 18.91}$ \\
llama-v3p1-70b & $28.15_{\pm 14.87}$ & $42.92_{\pm 17.82}$ & $27.63_{\pm 14.56}$ & $49.21_{\pm 19.09}$ & $50.79_{\pm 19.09}$ \\
\addlinespace

\textbf{N = 10} &&&&&\\
gpt-3.5-turbo-0125 & $44.06_{\pm 25.68}$ & $43.51_{\pm 25.37}$ & $41.55_{\pm 24.96}$ & $55.15_{\pm 27.01}$ & $44.85_{\pm 27.01}$ \\
gpt-4o-2024-05-13 & $43.33_{\pm 14.44}$ & $44.23_{\pm 13.8}$ & $40.92_{\pm 13.59}$ & $54.88_{\pm 16.33}$ & $45.12_{\pm 16.33}$ \\
llama-v3p1-70b & $38.49_{\pm 15.73}$ & $39.33_{\pm 16.04}$ & $36.82_{\pm 15.33}$ & $47.43_{\pm 17.39}$ & $52.57_{\pm 17.39}$ \\
\bottomrule
\end{tabular}
\end{table*}